\documentclass[10pt,twocolumn,letterpaper]{article}

\usepackage{cvpr}      

\definecolor{cvprblue}{rgb}{0.21,0.49,0.74}
\usepackage[pagebackref,breaklinks,colorlinks,allcolors=cvprblue]{hyperref}
\usepackage{booktabs}   
\usepackage{multirow}   
\usepackage{colortbl}   
\usepackage[table]{xcolor} 
\usepackage{graphicx}   
\usepackage{adjustbox}
\usepackage{algorithm}
\usepackage{algpseudocode}
\usepackage{amsmath}

\usepackage{booktabs} 
\usepackage{caption}  
\usepackage[table]{xcolor} 

\usepackage{listings}
\usepackage{tcolorbox}

\newtcolorbox{promptbox}[1][]{
  colback=gray!10!white,
  colframe=gray!50!black,
  title=\textbf{System Instruction},
  fonttitle=\bfseries,
  boxrule=0.5mm,
  arc=2mm,
  bottom=2pt,
  top=2pt,
  #1
}

\def\paperID{} 
\def\confName{CVPR}
\def\confYear{2026}

\title{CoAgent: Collaborative Planning and Consistency Agent for Coherent Video Generation}

\author{
\textbf{Qinglin Zeng}$^{1}$,
\textbf{Kaitong Cai}$^{1}$,
\textbf{Ruiqi Chen}$^{1}$,
\textbf{Qinhan Lv}$^{1}$,
\textbf{Keze Wang}$^{1,\dagger}$ \\
\vspace{0.5em}
$^{1}$Sun Yat-sen University \\
\vspace{0.3em}
$^{\dagger}$Corresponding author: \texttt{kezewang@gmail.com}
}

\begin{document}
\maketitle
\begin{abstract}
Maintaining narrative coherence and visual consistency remains a central challenge in open-domain video generation. Existing text-to-video models typically treat each shot independently, ignoring inter-shot dependencies and leading to identity drift, scene discontinuity, and unstable pacing. We present \textbf{CoAgent} (Collaborative Planning and Consistency Agent), a collaborative, closed-loop framework that formulates video synthesis as a plan–synthesize–verify–edit process. Given a user-defined prompt, style reference, and pacing template, a \emph{Storyboard Planner} decomposes abstract concepts into structured shot plans with explicit entities, spatial relations, and temporal cues. A \emph{Global Context Manager} (GCM) maintains entity-level memory to ensure consistent identities and appearances across shots. Each shot is rendered by a \emph{Synthesis Module} under the guidance of a \emph{Visual Consistency Controller}, while a \emph{Verifier Agent} evaluates intermediate frames using a vision–language model and triggers selective regeneration when inconsistencies are detected. Finally, a pacing-aware editor refines rhythm and transitions according to the target tempo. This agentic and modular design bridges creative planning with automated quality control, achieving coherent, style-aligned, and rhythm-consistent video generation.
\end{abstract}
    
\section{Introduction}
\label{sec:intro}
\begin{figure}[t]
    \centering
    \includegraphics[width=0.48\textwidth]{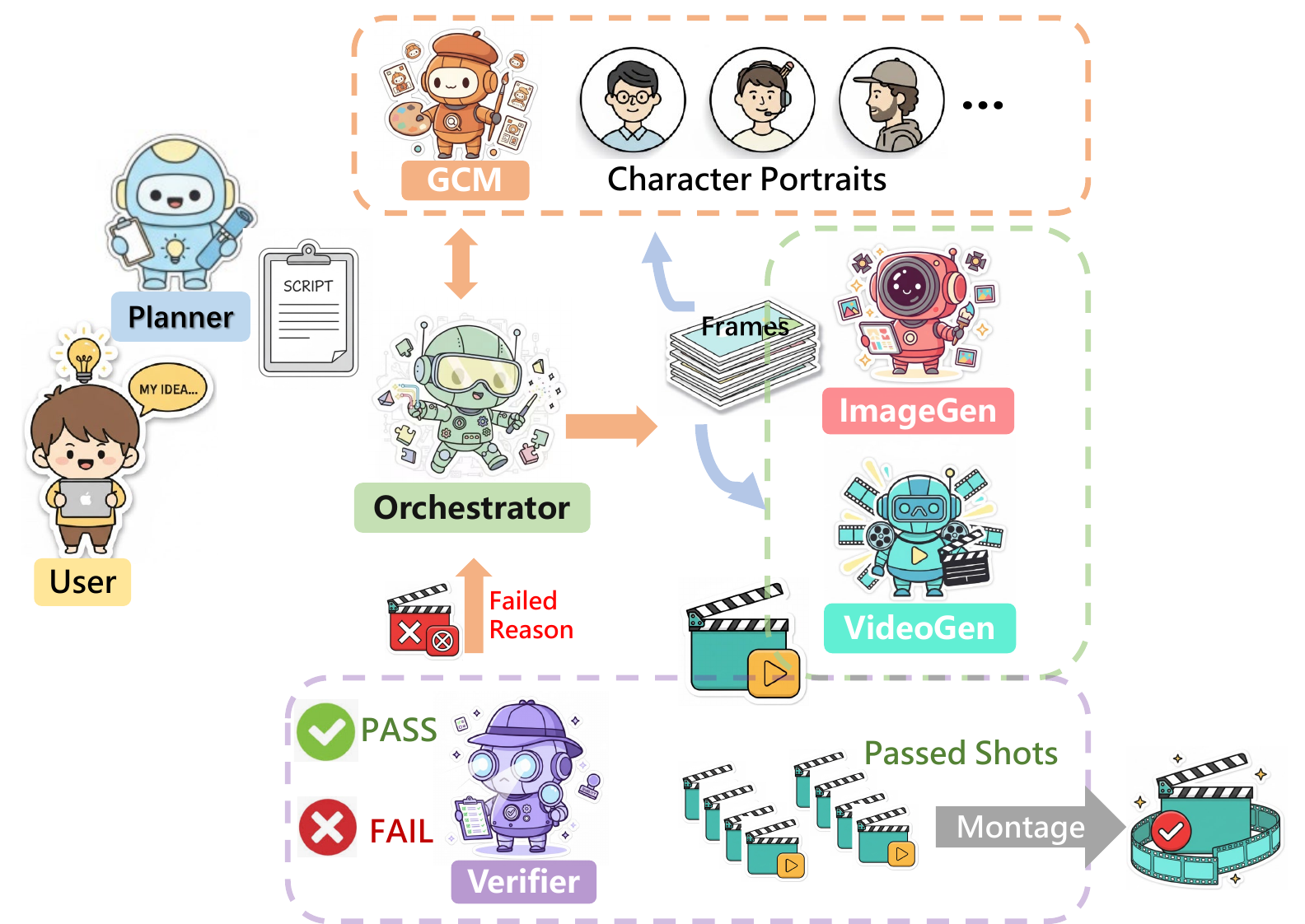}
    
    \caption{The overall architecture of our framework.}
    
    \label{fig:framework}
\end{figure}
Open-domain video generation requires not only visual realism within individual clips but also narrative coherence and cross-shot consistency throughout long sequences.  
While recent text-to-video (T2V) diffusion models~\cite{wan2025wan, wang2025cinemaster,z1,z2,z9} excel at generating short, semantically aligned clips, they remain fragile when tasked with producing extended multi-shot stories involving recurring characters, evolving scenes, and rhythmically structured pacing.  
The root of this limitation lies in how the generation process is formulated\cite{Z3,z4,dinkevich2025story2board,he-etal-2024-videoscore}.

Most existing pipelines treat video generation as a single-pass, open-loop translation from script to video.  
Given a script decomposed into $N$ shots, they independently generate each shot $i$ using a T2V model $\mathcal{F}$ with textual prompt $p_i$, which can be summarized as
$
V = \bigoplus_{i=1}^{N} \mathcal{F}(p_i),
$
where $\bigoplus$ denotes concatenation.  
This paradigm is inherently \textit{stateless}: the generation of $s_j$ is conditionally independent of $s_i$ for $i \neq j$.  
Without any persistent memory or feedback, such models cannot recall an entity’s exact visual identity from one shot and reproduce it consistently in subsequent scenes.  
Consequently, long-form videos suffer from \textbf{identity drift}, \textbf{scene discontinuity}, and unstable visual style—issues that break narrative immersion and limit creative control\cite{Z5,z6,cheng2025vpo,wan2025wan,z7,z22}.

We argue that coherent video generation should not be a static mapping $\mathcal{F}$ but a dynamic, feedback-driven reasoning process.  
To this end, we introduce \textbf{CoAgent}—a \textit{Collaborative Planning and Consistency Agent} that reformulates video synthesis as a closed-loop, multi-agent process of \textit{planning, generation, verification, and refinement}.  
Unlike prior prompt-based or globally conditioned approaches, CoAgent explicitly maintains state across shots through structured memory and inter-agent collaboration, enabling both continuity and controllability\cite{z8,dinkevich2025story2board,zheng2025vbench,z10}.

At the core of CoAgent lies a \emph{Storyboard Planner} ($\mathcal{A}_{plan}$) that decomposes a high-level user concept $P_{idea}$ into a structured shot plan $\mathcal{S} = \{S_1, \dots, S_N\}$ describing entities, spatial–temporal relations, and pacing intent.  
To preserve identities and context across shots, we introduce a \emph{Global Context Manager} (GCM), an explicit cross-shot memory $\mathcal{M}_{GCM}$ that registers visual representations of key entities $e_k$ such as characters or props.  
Conditioned on this memory, the \emph{Synthesis Module} ($\mathcal{A}_{synth}$) renders each shot as
$
s_i = \mathcal{A}_{synth}(S_i, \mathcal{M}_{GCM}, s_{i-1}),
$
retrieving entity appearance from $\mathcal{M}_{GCM}$ and optionally referencing the previous shot $s_{i-1}$—for example, under ff2v or flf2v modes—to maintain temporal smoothness under the guidance of a \emph{Visual Consistency Controller}.  
This conditioning transforms coherence from a fragile prompt heuristic into an explicit, state-aware mechanism.

To close the loop, a \emph{Verifier Agent} ($\mathcal{A}_{verify}$) employs a vision–language model to assess each synthesized shot and produces a verification signal
$\mathcal{V}_i = \mathcal{A}_{verify}(s_i, S_i, \mathcal{M}_{GCM}),
$quantifying both shot fidelity and cross-shot consistency.  
If $\mathcal{V}_i < \tau$, indicating inconsistency, CoAgent triggers selective regeneration—refining $S_i$ or adjusting the synthesis mode within $\mathcal{A}_{synth}$—thereby instituting a self-correcting feedback loop.  
A pacing-aware editor subsequently reconciles rhythm and transitions, ensuring that the final video aligns with the intended narrative tempo and mood.

In summary, \textbf{CoAgent} transforms video generation from a stateless, open-loop pipeline into a stateful, agentic framework.  
By coupling explicit global memory with verification-driven feedback, it shifts the responsibility for coherence from brittle prompt engineering to structured, collaborative reasoning.  
Our main contributions are summarized as follows:
\begin{enumerate}
    \item We propose \textbf{CoAgent}, a collaborative closed-loop paradigm that unifies planning, synthesis, and verification for multi-shot video generation.
    \item We introduce an explicit entity-level memory module $\mathcal{M}_{GCM}$ that preserves identity and appearance consistency across shots through structured retrieval and conditioning.
    \item We develop a verifier-guided adaptive synthesis strategy that achieves a controllable trade-off between generation efficiency and visual fidelity via selective regeneration.
\end{enumerate}
\section{Related Work}
\label{sec:related}

Our work, CoAgent, addresses the challenge of long-form, coherent video generation by integrating narrative planning, stateful memory, and closed-loop verification. Our contributions are thus situated at the intersection of three primary research thrusts: (1) spatiotemporal consistency preservation, (2) high-level narrative and compositional planning, and (3) agentic, feedback-driven generative systems.

\subsection{Passive Visual State Memory and Consistency}
A significant body of work focuses on mitigating the "visual failure" component of the narrative gap, such as identity drift and scene incoherence. 

\textbf{Identity Preservation.} Maintaining character identity is a critical sub-problem. A popular approach involves using tuning-free adapters~\cite{ye2023ip,wang2024instantid,li2024photomaker,z11,z12} to inject identity features, extracted from reference images via encoders like CLIP or ArcFace, into the cross-attention layers of pre-trained T2V models. Other methods propose architectural modifications or specialized training objectives. For example, ConsisID~\cite{yuan2025identity,z13} utilizes a frequency-aware heuristic in a DiT model to preserve facial information, while MotionCharacter~\cite{fang2024motioncharacter,z14,z15} employs a composite loss, including an ID-consistency term, to maintain identity during complex motions.

\textbf{Explicit Memory Architectures.} To address consistency over longer horizons, more advanced frameworks have introduced explicit memory mechanisms. These methods recognize that long-term coherence requires a persistent state. For instance, Corgi~\cite{wu2025corgi} introduces a "cached memory mechanism" that stores keyframes of previously generated scenes in a latent memory bank, conditioning new scene generation on this cache. Similarly, SPMem~\cite{wu2025video} augments a model's working memory with a geometry-grounded 3D point map to preserve static scene information.

While these methods effectively enhance visual statefulness, their memory is often \textit{passive}. They can recall what an entity looked like, but they lack high-level semantic planning to determine what the entity should do next according to a narrative.

\subsection{Blind Execution of Open-Loop Planning}
This second research thrust tackles the "semantic failure" component by introducing high-level intent into the generation process, typically using Large Language Models (LLMs) as planners. These frameworks often follow a "plan-then-generate" paradigm.

VGoT~\cite{zheng2024videogen}, for example, employs an LLM to decompose a high-level prompt into a detailed, multi-shot storyline with cinematic specifications. This plan is then executed by a T2V model in a sequential, open-loop fashion, using "identity-aware cross-shot propagation" tokens to pass identity information. Similarly, Story2Board~\cite{dinkevich2025story2board} uses an LLM "Director" to generate grounded panel-level prompts for storyboard synthesis, employing a "Latent Panel Anchoring" technique to enforce identity during a batch denoising process.

The fundamental limitation of these approaches is their \textbf{open-loop} nature. The generation stage is a passive execution of the pre-defined plan. If the synthesis module produces an error or deviates from the plan (e.g., incorrect object, failed action), the system has no feedback mechanism to detect this failure or dynamically re-plan. This can lead to the very error accumulation~\cite{meng2025holocine,z16,z17,z18} they were designed to prevent.

\subsection{Limited Closed-Loop Systems }
To overcome the brittleness of open-loop planning, an emerging paradigm reformulates synthesis as a dynamic, closed-loop, and collaborative process. This approach often involves multi-agent systems (MAS) and a verification mechanism.

\textbf{Iterative Refinement.} The core concept of a verification-driven loop was notably demonstrated for static images by SLD~\cite{wu2024self}. SLD employs a "generate-assess-correct" cycle where an LLM controller assesses a generated image for mismatches against the prompt and suggests latent-space corrections.

\textbf{Multi-Agent Video Synthesis.} This concept has been extended to video. Kubrick~\cite{he2024kubrick,z19,z20} uses a multi-agent system (Director, Programmer, Reviewer) to automate 3D production. It implements a closed-loop process where the Reviewer provides feedback to the Programmer. However, it relies on a deterministic 3D rendering engine (e.g., Blender), bypassing the stochastic challenges of end-to-end diffusion models. Closer to our domain, GenMAC~\cite{huang2024genmac} proposes a multi-agent framework for compositional T2V generation with an iterative "redesign" loop, allowing for the correction of compositional errors. Hollywood Town~\cite{wei2025hollywood} introduces a graph-based multi-agent framework with a "controlled cyclic execution strategy" that allows agents to reflect and refine outputs.

\noindent\textbf{Position of CoAgent.}
While these agentic frameworks pioneer the closed-loop approach, they often focus on compositional correctness (GenMAC) or graph-based execution (Hollywood Town) and typically lack a persistent, stateful memory dedicated to narrative entities.

To date, these three research thrusts—(1) spatiotemporal memory, (2) high-level planning, and (3) closed-loop verification—have remained largely separate, leaving a critical gap. The key scientific challenge is not to simply combine them, but to solve the \textbf{non-trivial integration problem}: How does a planner \emph{write to} a memory? How does a verifier \emph{read from} that memory? And how does verification failure trigger a \emph{dynamic re-plan} rather than a simple failure?

Our work, \textbf{CoAgent}, is the first to be proposed to solve this synergistic challenge. Our novelty lies in the design of the interfaces that unify these three components:
\begin{itemize}
    \item[(1)] \textbf{Dynamic Planning:} We \emph{transform} the static "plan-then-generate" paradigm of VGoT~\cite{zheng2024videogen} into a dynamic, state-aware process.
    
    \item[(2)] \textbf{Active Memory:} We \emph{introduce} an \textbf{active memory} (our GCM), which, unlike the passive caches of Corgi~\cite{wu2025corgi}, is explicitly queried by the \emph{Planner} and updated by the \emph{Verifier}.
    
    \item[(3)] \textbf{Narrative-Aware Verification:} We \emph{propose} a \textbf{narrative-aware verification} loop. Instead of just assessing static images (like SLD~\cite{wu2024self}), our Verifier validates multi-shot narrative logic and cross-shot entity consistency against the GCM, enabling it to detect complex failures.
\end{itemize}
To our knowledge, CoAgent is the first framework to integrate these three concepts into a single, stateful, and truly agentic system for coherent video generation.

\section{Method}
\label{sec:method}
\begin{figure*}[t]
    \centering
    \includegraphics[width=0.90\textwidth]{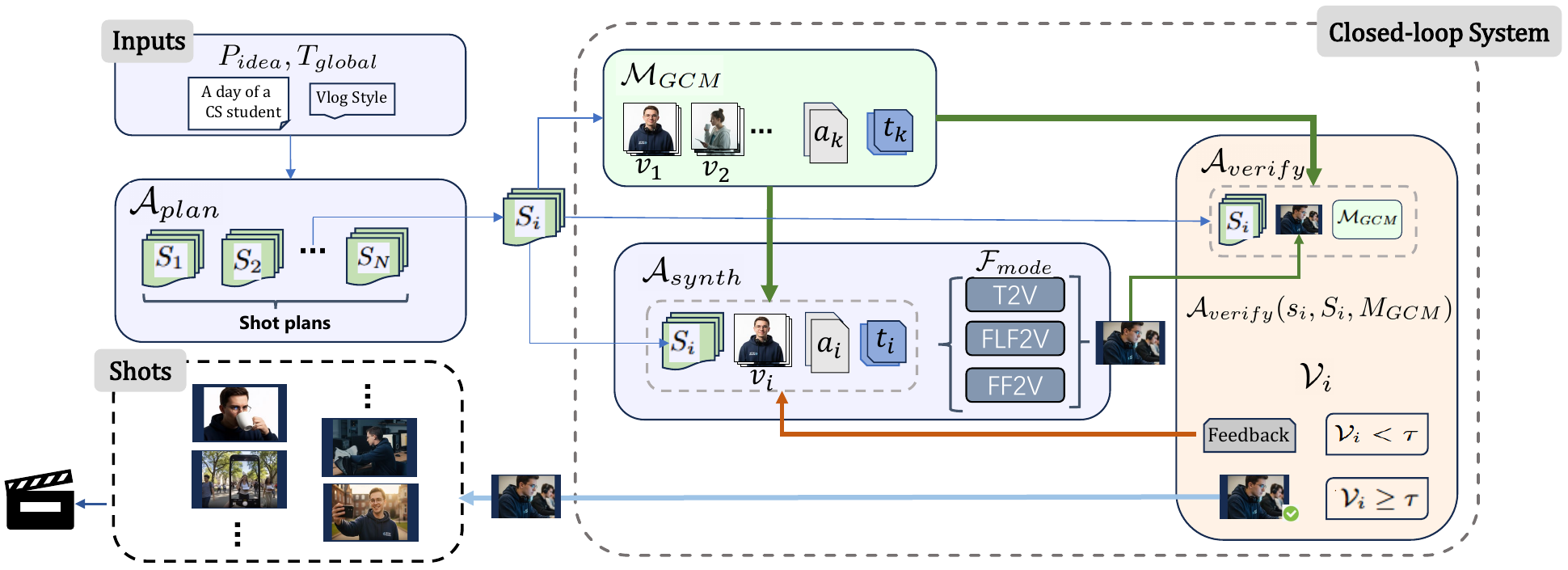}
    
    \caption{
        The detailed architecture of our CoAgent workflow. 
        The framework integrates four key modules: 
        the \textbf{Storyboard Planner} ($\mathcal{A}_{plan}$), 
        the \textbf{Global Context Manager} ($\mathcal{M}_{GCM}$), 
        the \textbf{Synthesis Module} ($\mathcal{A}_{synth}$), 
        and the \textbf{Verifier Agent} ($\mathcal{A}_{verify}$).
    }
    
    \label{fig:workflow}
\end{figure*}
\subsection{Overview of the Framework}
\label{sec:method:overview}

Following the motivation and problem formulation discussed in Section~\ref{sec:intro},  
we present the proposed \textbf{CoAgent} framework—our core contribution that formulates open-domain video generation as a closed-loop, multi-agent process integrating \emph{planning}, \emph{generation}, \emph{verification}, and \emph{refinement}.  
Unlike conventional open-loop pipelines, CoAgent establishes a stateful generation mechanism grounded in explicit memory and collaborative reasoning among agents.  
Figure~\ref{fig:workflow} illustrates the overall architecture, while Algorithm~\ref{alg:coagent} and Algorithm~\ref{alg:verifyloop} detail the high-level workflow and the verifier-guided regeneration subroutine.

Given a high-level concept $P_{idea}$ and optional pacing or style templates, CoAgent generates a coherent video $V = \{s_1, \dots, s_N\}$ through four interdependent modules:
\textbf{Storyboard Planner} $\mathcal{A}_{plan}$ decomposes $P_{idea}$ into structured shot plans $\mathcal{S} = \{S_i\}_{i=1}^N$;
\textbf{Global Context Manager (GCM)} maintains entity-level representations within a cross-shot memory $\mathcal{M}_{GCM}$;
\textbf{Synthesis Module} $\mathcal{A}_{synth}$ generates each shot $s_i$ conditioned on $(S_i, \mathcal{M}_{GCM}, s_{i-1})$;
\textbf{Verifier Agent} $\mathcal{A}_{verify}$ evaluates generated results and triggers regeneration when inconsistencies arise.
These modules interact through a recurrent plan–synthesize–verify–edit loop.  
The closed-loop process, described in Algorithm~\ref{alg:coagent}, transforms traditional open-loop generation into a controllable, feedback-driven reasoning paradigm.

\subsection{Storyboard Planning}
\label{sec:method:planning}
The planner $\mathcal{A}_{plan}$ converts $P_{idea}$ into a structured sequence of shot descriptors $\mathcal{S} = \{S_i\}$, where each shot is parameterized as:
\begin{equation}
S_i = (p_i, E_i, R_i, \text{style}_i, \text{duration}_i).
\label{eq:shot}
\end{equation}
Here $p_i$ is the textual prompt, $E_i = \{e_k\}$ represents key entities, $R_i$ defines spatial–temporal relations, and $(\text{style}_i, \text{duration}_i)$ specify local rhythm and tone.  
Global indexing ensures consistent entity references across shots.  
This stage establishes a bijective mapping between high-level semantics and structured visual primitives, enabling differentiable control for downstream synthesis.

\subsection{Global Context Manager (GCM)}
\label{sec:method:gcm}

The GCM functions as a persistent cross-shot memory $\mathcal{M}_{GCM}$, storing canonical entity embeddings for retrieval and update:
\begin{equation}
\mathcal{M}_{GCM}(e_k) = (\mathbf{v}_k, \mathbf{a}_k, t_k),
\label{eq:gcm}
\end{equation}
where $\mathbf{v}_k$ encodes visual appearance, $\mathbf{a}_k$ represents auxiliary attributes (pose, emotion, lighting), and $t_k$ records the last update step.  
When an entity reappears, $\mathbf{v}_k$ is retrieved and fused via cross-attention into the synthesis process.  
This formulation converts inter-shot coherence modeling from pairwise comparison to single-pass retrieval, reducing complexity from $O(N^2)$ to $O(N)$ and enabling scalable memory-based consistency.

\begin{algorithm}[t]
\caption{CoAgent Closed-Loop Video Generation}
\label{alg:coagent}
\begin{algorithmic}[1]
\Require Concept $P_{idea}$, pacing template $T_{global}$
\State $\mathcal{S} \gets \mathcal{A}_{plan}(P_{idea})$ \Comment{Storyboard planning}
\State Initialize $\mathcal{M}_{GCM} \gets \emptyset$
\For{$i = 1$ to $N$}
    \State $S_i \gets \mathcal{S}[i]$
    \State Select mode $\mathcal{F}_{mode}$ based on $\|\mathcal{V}_{i-1}-\tau\|$ and entity overlap
    \State $s_i \gets \mathcal{A}_{synth}(S_i, \mathcal{M}_{GCM}, s_{i-1})$
    \State $[s_i, \mathcal{M}_{GCM}] \gets \text{VerifierLoop}(S_i, s_i, \mathcal{M}_{GCM})$
\EndFor
\State $V \gets \text{PacingEditor}(\{s_i\}, T_{global})$
\State \Return Final coherent video $V$
\end{algorithmic}
\end{algorithm}

\subsection{Shot Synthesis Module}
\label{sec:method:synth}

Given a shot plan $S_i$ and global memory $\mathcal{M}_{GCM}$, the synthesis agent $\mathcal{A}_{synth}$ generates the visual sequence
\begin{equation}
s_i = \mathcal{A}_{synth}(S_i, \mathcal{M}_{GCM}, s_{i-1}),
\label{eq:synth}
\end{equation}
where $s_{i-1}$ provides temporal context under a \textbf{Visual Consistency Controller (VCC)}.  
CoAgent adaptively selects one of three generation modes $\mathcal{F}_{mode}$:
\textbf{T2V:} $s_i = \mathcal{F}_{t2v}(S_i, \mathcal{M}_{GCM})$, used for initial or isolated shots;
\textbf{FF2V:} $s_i = \mathcal{F}_{ff2v}(S_i, \mathcal{M}_{GCM}, \text{last}(s_{i-1}))$, anchoring the first frame of $s_i$ to the last frame of $s_{i-1}$;
\textbf{FLF2V:} $s_i = \mathcal{F}_{flf2v}(S_i, \mathcal{M}_{GCM}, \text{last}(s_{i-1}), I_{goal})$, using both starting and goal frames for bidirectional continuity.
Mode selection is adaptively governed by $\|\mathcal{V}_{i-1}-\tau\|$ and the entity overlap between $E_{i-1}$ and $E_i$, enabling progressive temporal anchoring.  
The VCC computes feature-space similarity between retrieved $\mathbf{v}_k$ and generated frames to modulate diffusion guidance and maintain stylistic and motion consistency.
\subsection{Verifier and Feedback Loop}
\label{sec:method:verify}

After synthesis, the Verifier Agent $\mathcal{A}_{verify}$ is implemented as
a frozen VLM prompted to critique each shot. Given the generated shot $s_i$,
its storyboard specification $S_i$, and the global memory $\mathcal{M}_{GCM}$,
the verifier produces \emph{structured feedback} and a scalar quality score:
\begin{equation}
(\text{fb}_i, \mathcal{V}_i) = \mathcal{A}_{verify}(s_i, S_i, \mathcal{M}_{GCM}),
\label{eq:verify}
\end{equation}
where $\text{fb}_i$ describes the error type (e.g., semantic mismatch, missing
entities, temporal discontinuity) and suggested corrections, and
$\mathcal{V}_i \in [0,1]$ summarizes the overall visio-semantic quality.

When $\mathcal{V}_i < \tau$, CoAgent activates a verifier-guided regeneration
routine (Algorithm~\ref{alg:verifyloop}). The textual feedback $\text{fb}_i$
is fed back to the synthesis pipeline to decide \emph{how} to modify the
storyboard or synthesis mode, while the scalar $\mathcal{V}_i$ serves as a
stopping criterion: the loop continues until the verified quality of each shot
exceeds the threshold.
\begin{algorithm}[t]
\caption{Verifier-Guided Regeneration and Memory Update}
\label{alg:verifyloop}
\begin{algorithmic}[1]
\Require Shot description $S_i$, generated shot $s_i$, memory $\mathcal{M}_{GCM}$, threshold $\tau$
\State $(\text{fb}_i, \mathcal{V}_i) \gets \mathcal{A}_{verify}(s_i, S_i, \mathcal{M}_{GCM})$ 
\While{$\mathcal{V}_i < \tau$}
    \If{\text{fb$_i$ indicates semantic misalignment}}
        \State $S_i \leftarrow \text{RefineSemantic}(S_i, \text{fb}_i)$
    \ElsIf{\text{fb$_i$ indicates appearance inconsistency}}
        \State Switch synthesis mode (T2V$\!\rightarrow$FF2V or FF2V$\!\rightarrow$FLF2V)
    \EndIf
    \State $s_i' \gets \mathcal{A}_{synth}(S_i, \mathcal{M}_{GCM}, s_{i-1})$
    \State $(\text{fb}_i, \mathcal{V}_i) \gets \mathcal{A}_{verify}(s_i', S_i, \mathcal{M}_{GCM})$
\EndWhile
\State Update $\mathcal{M}_{GCM}$ with embeddings from $s_i'$
\State \Return Verified $s_i'$, updated $\mathcal{M}_{GCM}$
\end{algorithmic}
\end{algorithm}


\section{Experiments}
\label{sec:experiments}

To validate the effectiveness of our \textbf{CoAgent} framework, we conduct a comprehensive set of experiments. Our evaluation is designed to answer three key questions:
(1) Does our collaborative, closed-loop framework outperform state-of-the-art (SOTA) open-loop and agent-based methods in narrative coherence and visual consistency?
(2) What is the specific contribution of each core component, namely the \emph{Storyboard Planner}, the \emph{Global Context Manager (GCM)}, and the \emph{Verifier Agent}?
(3) How does our framework enhance a powerful foundation model, and how does it compare against other leading closed-source models?

\subsection{Experimental Setup}
\label{sec:setup}

\noindent\textbf{Evaluation Benchmarks.}
We employ two state-of-the-art benchmarks for automatic video evaluation, chosen for their comprehensive, multi-dimensional assessment and high correlation with human judgment.
\begin{itemize}
    \item \textbf{VBench}~\cite{zheng2025vbench}: A comprehensive benchmark suite that decomposes video quality into multiple fine-grained dimensions. We report results on seven key metrics: \emph{Subject Consistency}, \emph{Background Consistency}, \emph{Temporal Flickering}, \emph{Motion Smoothness}, \emph{Aesthetic Quality}, \emph{Imaging Quality}, and \emph{Temporal Style}.
    \item \textbf{VideoScore}~\cite{he-etal-2024-videoscore}: A metric trained on \emph{VideoFeedback}, a large-scale dataset of human multi-aspect ratings. We use VideoScore-v1.1 to evaluate \emph{Visual Quality}, \emph{Temporal Consistency}, \emph{Dynamic Degree}, \emph{Text-Video Alignment}, and \emph{Factual Consistency}.
\end{itemize}

\noindent\textbf{Baselines.}
We compare CoAgent against two categories of SOTA methods:
\begin{enumerate}
    \item \textbf{Agent-based and Prompt Optimization Methods:} We compare against the baselines reported in VISTA~\cite{long2025vista}, a SOTA multi-agent system for video generation. These include: \textbf{DP} (Direct Prompting), \textbf{VSR}~\cite{madaan2023self} and \textbf{VSR++} (Visual Self-Refine), \textbf{Rewrite} (LLM-based prompt rewriting), \textbf{VPO}~\cite{cheng2025vpo} (prompt expansion), and \textbf{VISTA} itself.
    \item \textbf{Foundation Models:} We compare against leading open-source and closed-source video generation models: \textbf{CogVideoX}~\cite{yang2024cogvideox}, \textbf{OpenSoraV2}~\cite{peng2025open}, \textbf{Wan2.1}~\cite{wan2025wan}, and \textbf{Sora2}~\cite{openai_sora2_2025}.
\end{enumerate}

\noindent\textbf{Implementation Details.}
Our CoAgent framework orchestrates several models. The \emph{Storyboard Planner} ($\mathcal{A}_{plan}$) is powered by Gemini2.5-Flash. The core \emph{Synthesis Module} ($\mathcal{A}_{synth}$) uses the pre-trained \textbf{Wan2.1} model as its backbone. The \emph{Verifier Agent} ($\mathcal{A}_{verify}$) and \emph{Global Context Manager} (GCM) utilize GPT-4o, supplemented by Gemini2.5-Flash-Image for specific visual processing tasks. For evaluations involving the closed-source Sora2 model, videos were generated via its web interface, and the platform's dynamic watermark was removed prior to evaluation to ensure a fair comparison. All experiments were conducted using prompts sampled from the VBench dataset and other standard prompt lists. Further details of the experiments are provided in the \textbf{Supplementary Material (Sec. A)}.

\subsection{Comparison with State-of-the-Art}
\label{sec:sota_comparison}

\subsubsection{Comparison with Agent-based Frameworks}
We first evaluate CoAgent against VISTA and its associated baselines on the VBench benchmark. The results, presented in Table~\ref{tab:vbench_results}, demonstrate the superiority of our framework.

CoAgent outperforms all competing methods across all seven evaluation dimensions. The most significant gains are in consistency metrics: we achieve a score of \textbf{94.70} in \emph{Subject Consistency} and \textbf{96.50} in \emph{Background Consistency}, surpassing the next-best method (VISTA) by 4.75 and 3.61 points, respectively. This highlights the substantial advantage of our \emph{Global Context Manager (GCM)}. Unlike VISTA, which relies on indirect prompt refinement to maintain coherence, our GCM maintains an explicit entity-level memory, enabling robust identity and appearance preservation across shots.

Furthermore, CoAgent achieves the highest scores in \emph{Temporal Flickering} (99.20) and \emph{Motion Smoothness} (99.40). This validates the effectiveness of our \emph{Verifier Agent}. By operating in a closed-loop, the verifier identifies and triggers the regeneration of frames with temporal artifacts, leading to a more stable and visually pleasing video output.
\begin{table*}[t]
\centering
\caption{Quantitative comparison with agent-based SOTA methods on \textbf{VBench}. All scores are percentages ($\uparrow$); higher is better. Our CoAgent framework significantly outperforms all prompt-optimization and agent-based baselines, especially in subject and background consistency, demonstrating the effectiveness of our GCM and Verifier agent.}
\label{tab:vbench_results}
\definecolor{lightpink}{HTML}{F8D0D8}
\definecolor{lightblue}{HTML}{D0E0F8}
\rowcolors{2}{lightblue!50}{white} 
\begin{tabular}{lccccccc}
\toprule
\textbf{Metric} & \textbf{DP} & \textbf{VSR} & \textbf{VSR++} & \textbf{Rewrite} & \textbf{VPO} & \textbf{VISTA} & \textbf{CoAgent (Ours)} \\
\midrule
Subject Consistency & 89.89 & 89.33 & 87.96 & 89.09 & 86.74 & 89.95 & \textbf{94.70} \\
Background Consistency & 94.39 & 93.53 & 93.53 & 93.79 & 92.66 & 92.89 & \textbf{96.50} \\
Temporal Flickering & 97.82 & 97.79 & 97.88 & 97.59 & 97.76 & 97.82 & \textbf{99.20} \\
Motion Smoothness & 99.23 & 99.26 & 99.12 & 99.17 & 99.15 & 98.94 & \textbf{99.40} \\
Aesthetic Quality & 61.86 & 63.45 & 60.68 & 62.52 & 61.17 & 64.53 & \textbf{64.60} \\
Imaging Quality & 64.42 & 65.53 & 63.06 & 62.58 & 64.01 & 65.89 & \textbf{66.50} \\
Temporal Style & 7.88 & 9.26 & 9.25 & 8.57 & 8.03 & 9.63 & \textbf{9.70} \\
\bottomrule
\end{tabular}
\end{table*}
\subsubsection{Comparison with Foundation Models}
To assess our framework's ability to enhance a powerful, pre-trained generator, we use VideoScore~\cite{he-etal-2024-videoscore} to compare our full system against its own backbone (Wan2.1) and other leading foundation models.

As shown in Table~\ref{tab:videoscore_results}, the CoAgent framework provides a massive boost to its Wan2.1 backbone. By integrating our agentic planning, memory, and verification, we improve \emph{Visual Quality} from 2.788 to \textbf{2.953} (+5.9\%) and \emph{Text-Video Alignment} from 2.550 to \textbf{2.731} (+7.1\%). This clearly demonstrates that our framework (labeled as CoAgent) is not just a simple wrapper but a powerful system that fundamentally enhances the generative capabilities of its underlying synthesizer.

When compared to all models, CoAgent achieves the highest scores in \emph{Visual Quality}, \emph{Temporal Consistency}, \emph{Text-Video Alignment}, and \emph{Factual Consistency}. Notably, we surpass the strong, closed-source Sora2 model in these critical narrative-focused dimensions. While Sora2 exhibits a higher \emph{Dynamic Degree}, our results confirm that CoAgent excels at its primary goal: generating high-fidelity, coherent, and semantically-aligned video stories.

\begin{table}[t]
\centering
\caption{Quantitative comparison with SOTA foundation models on \textbf{VideoScore v1.1}. All scores are ($\uparrow$); higher is better. Our framework (CoAgent) significantly enhances its Wan2.1 backbone and outperforms all other models, including Sora2, on narrative-critical metrics.}
\label{tab:videoscore_results}
\definecolor{lightpink}{HTML}{F8D0D8}
\definecolor{lightblue}{HTML}{D0E0F8}
\rowcolors{2}{lightblue!50}{white} 
\resizebox{\columnwidth}{!}{%
\begin{tabular}{lccccc}
\toprule
\textbf{Metric} & \textbf{CogVideoX} & \textbf{OpenSoraV2} & \textbf{wan2.1} & \textbf{Sora2} & \textbf{CoAgent (Ours)} \\
\midrule
Visual Quality & 2.614 & 2.750 & 2.788 & 2.734 & \textbf{2.953} \\
Temporal Consistency & 2.652 & 2.537 & 2.800 & 2.484 & \textbf{2.808} \\
Dynamic Degree & 2.643 & 2.771 & 2.517 & \textbf{2.984} & 2.847 \\
Text-Video Alignment & 2.310 & 2.524 & 2.550 & 2.469 & \textbf{2.731} \\
Factual Consistency & 2.610 & 2.415 & 2.642 & 2.172 & \textbf{2.770} \\
\bottomrule
\end{tabular}%
}
\end{table}

\subsection{Ablation Study}
\label{sec:ablation}

To precisely quantify the contribution of our key architectural components, we conduct a rigorous ablation study. We analyze three configurations, starting from a strong open-loop baseline and incrementally adding our modules. The results are presented in Table~\ref{tab:ablation_study}.

\begin{itemize}
    \item \textbf{(A) Baseline:} This model consists of the \emph{Storyboard Planner} ($\mathcal{A}_{plan}$) and the \emph{Synthesis Module} ($\mathcal{A}_{synth}$). It represents a powerful "plan-then-generate" open-loop system. As shown in Table~\ref{tab:ablation_study}, this baseline achieves a reasonable \emph{Text-Video Alignment} score (1.695) but struggles with visual consistency, scoring 90.60 on \emph{Subject Consistency} and 93.50 on \emph{Temporal Flickering}.
    
    \item \textbf{(B) + GCM:} We add the \emph{Global Context Manager} (GCM) to the baseline, creating an open-loop system with explicit cross-shot memory. The impact is immediate and precisely targeted: \emph{Subject Consistency} jumps by \textbf{+3.9} points to \textbf{94.50}, and \emph{Background Consistency} improves to \textbf{96.40}. This clearly isolates the GCM's role as a highly effective module for enforcing narrative entity consistency. We also observe marginal improvements in other metrics (e.g., \emph{Text-Video Alignment} to 1.778), which we attribute to the more stable visual context provided by the GCM for the synthesizer.
    
    \item \textbf{(C) CoAgent (Full):} We add the \emph{Verifier Agent} ($\mathcal{A}_{verify}$) to complete our full closed-loop framework. This component yields a dramatic improvement in all remaining metrics, operating on the more consistent state provided by (B). The \emph{Verifier}'s feedback loop significantly boosts temporal quality, with \emph{Temporal Flickering} rising from 94.70 to \textbf{99.20} (\textbf{+4.5} points) and \emph{Motion Smoothness} increasing from 95.10 to \textbf{99.40} (\textbf{+4.3} points). Simultaneously, it enforces high-level semantic fidelity, causing \emph{Text-Video Alignment} to skyrocket from 1.778 to \textbf{2.731} (a \text{53.6\%} relative increase).
\end{itemize}

This ablation study provides clear evidence for our design. The GCM and the Verifier are not redundant; they are complementary modules that each solve a distinct and critical failure mode of open-loop video generation. The GCM solves cross-shot \emph{visual memory}, while the Verifier solves \emph{temporal quality} and \emph{semantic alignment} via a closed-loop feedback mechanism. Further ablation studies, including component analysis and backbone generalization, are provided in the \textbf{Supplementary Material (Sec. B)}.

\begin{table*}[t!] 
\centering
\caption{
    \textbf{Ablation study of CoAgent's components.} 
    We start with a strong open-loop \textbf{Baseline (A)} (Planner + Synthesizer) and incrementally add our modules. 
    The results validate the distinct contribution of each component: the \textbf{GCM (B)} for consistency, and the \textbf{Verifier (C)} for temporal quality and semantic alignment. 
    VBench scores ($\uparrow$) are percentages; Text-Video Alignment ($\uparrow$) is from VideoScore.
}
\label{tab:ablation_study}
\resizebox{\textwidth}{!}{
\begin{tabular}{lccc|ccccc}
\toprule
& \multicolumn{3}{c|}{\textbf{Components}} & \multicolumn{2}{c}{\textbf{GCM Metrics (Consistency)}} & \multicolumn{3}{c}{\textbf{Verifier Metrics (Quality \& Semantics)}} \\
\textbf{Model} & $\mathcal{A}_{plan}$ & GCM & $\mathcal{A}_{verify}$ & Subj. Cons. $\uparrow$ & Bg. Cons. $\uparrow$ & Temp. Flick. $\uparrow$ & Motion Smooth. $\uparrow$ & Text Align. $\uparrow$ \\
\midrule
(A) Baseline & \checkmark & $\times$ & $\times$ & 90.60 & 95.50 & 93.50 & 94.80 & 1.695 \\
(B) + GCM & \checkmark & \checkmark & $\times$ & \textbf{94.50} & \textbf{96.40} & 94.70 & 95.10 & 1.778 \\
(C) CoAgent & \checkmark & \checkmark & \checkmark & 94.70 & 96.50 & \textbf{99.20} & \textbf{99.40} & \textbf{2.731} \\
\bottomrule
\end{tabular}%
}
\end{table*}

\subsection{Qualitative Analysis and Case Studies}
\label{sec:qualitative_analysis}
\begin{figure*}[t!]
    \centering
    
    \makebox[0.95\textwidth][s]{
        \parbox{0.1\textwidth}{\small \textbf{Baseline} \\ (No GCM)} \hfill
        \includegraphics[width=0.22\textwidth]{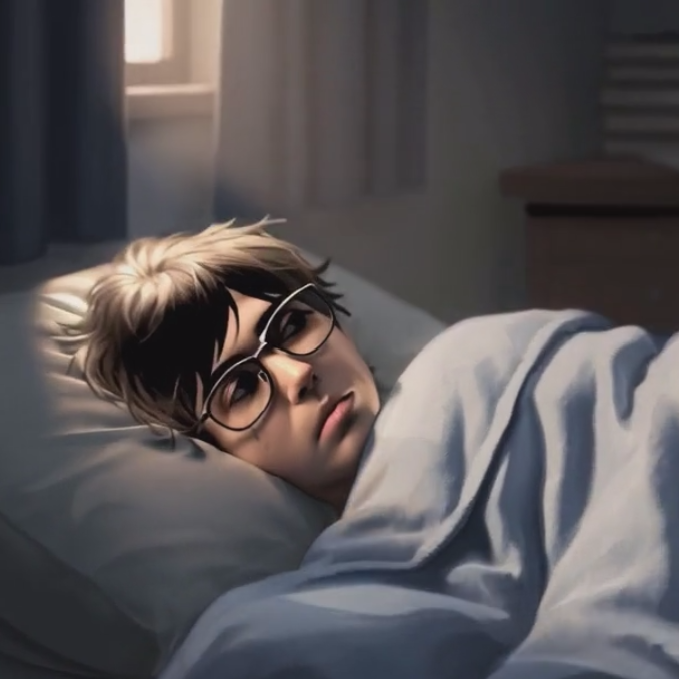} \hfill
        \includegraphics[width=0.22\textwidth]{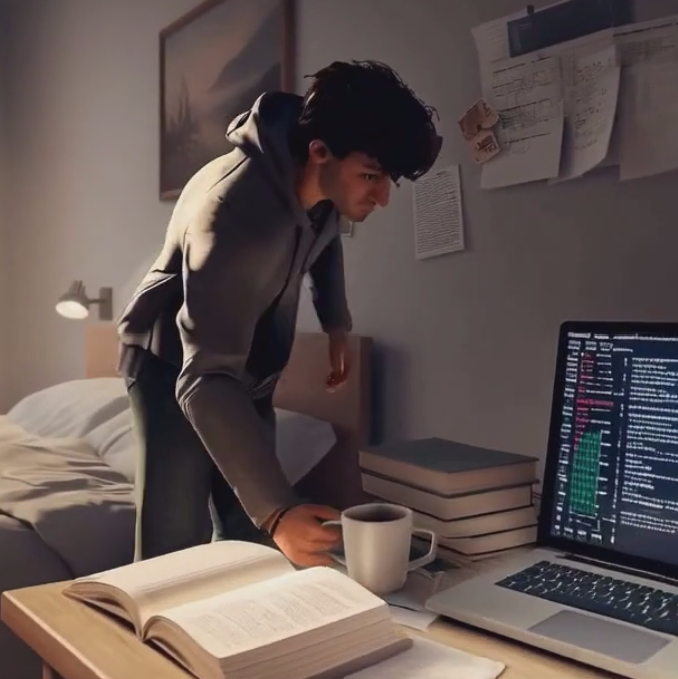} \hfill
        \includegraphics[width=0.22\textwidth]{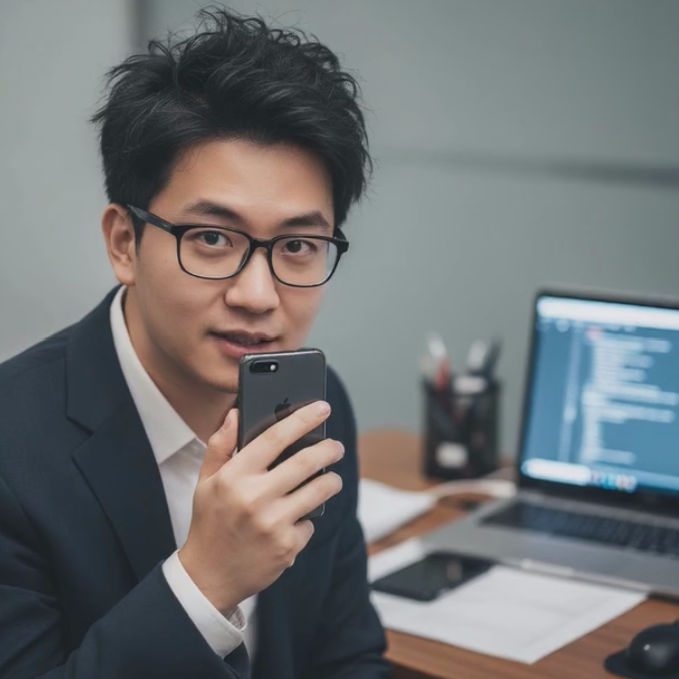} \hfill
        \includegraphics[width=0.22\textwidth]{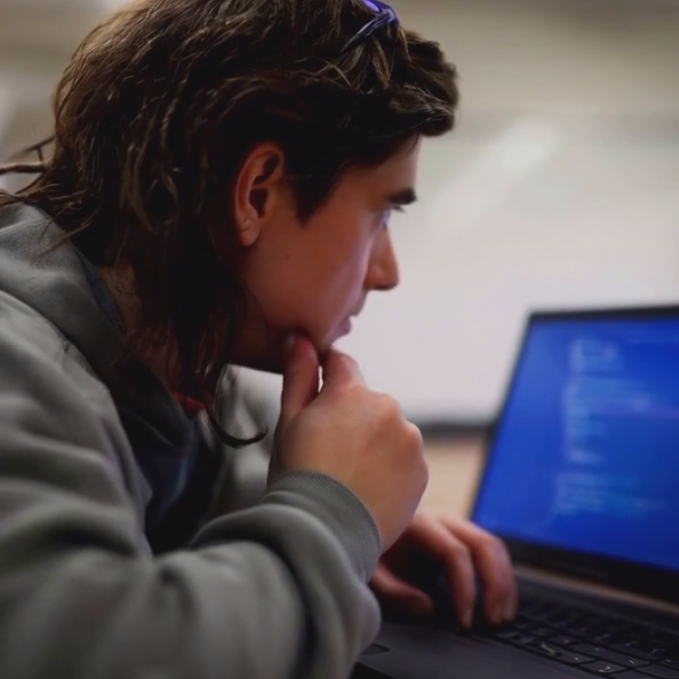}
    }
    \vspace{1mm} 
    \makebox[0.95\textwidth][s]{
        \parbox{0.1\textwidth}{\small \textbf{CoAgent (Ours)} \\ (with GCM)} \hfill
        \includegraphics[width=0.22\textwidth]{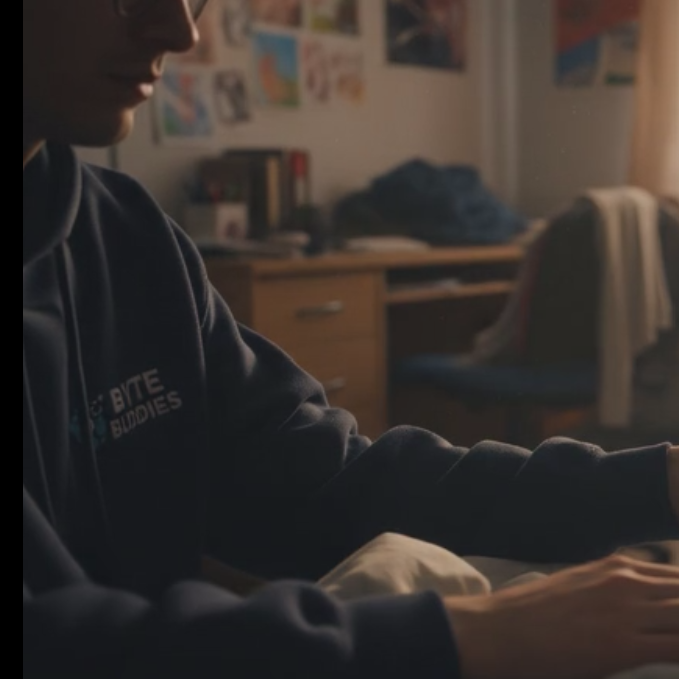} \hfill
        \includegraphics[width=0.22\textwidth]{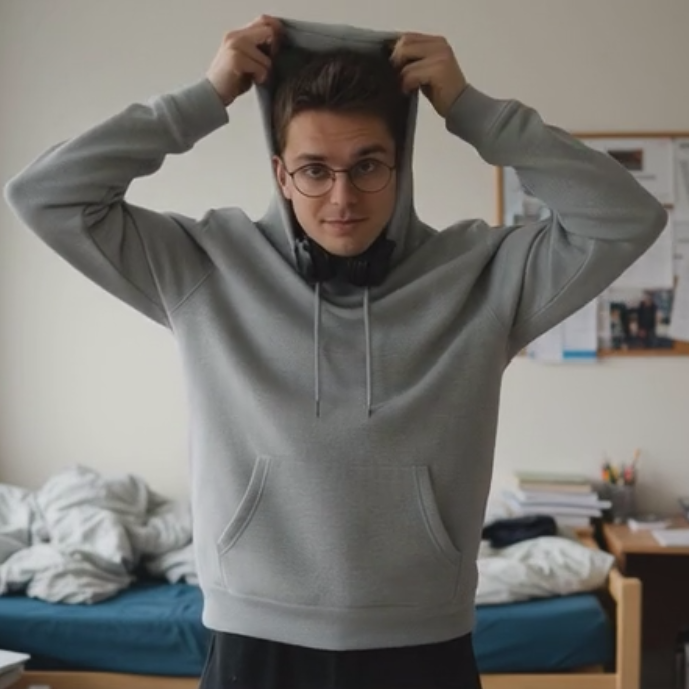} \hfill
        \includegraphics[width=0.22\textwidth]{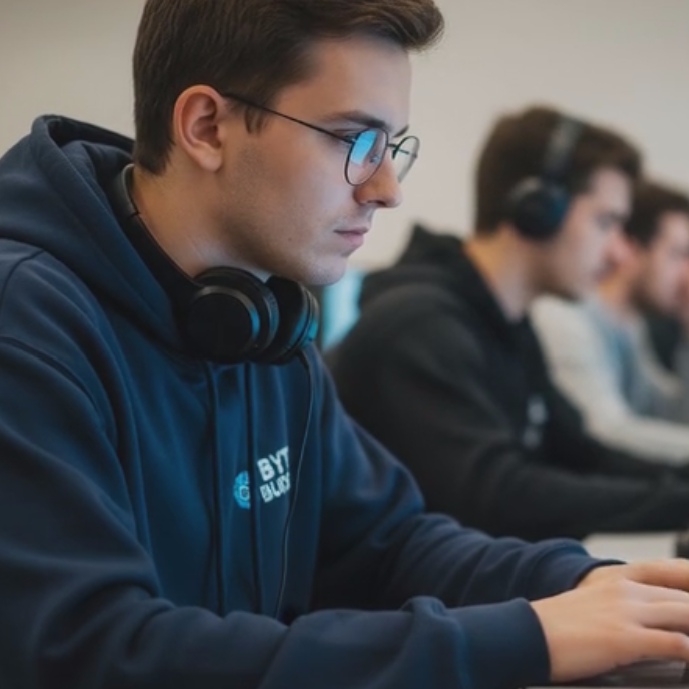} \hfill
        \includegraphics[width=0.22\textwidth]{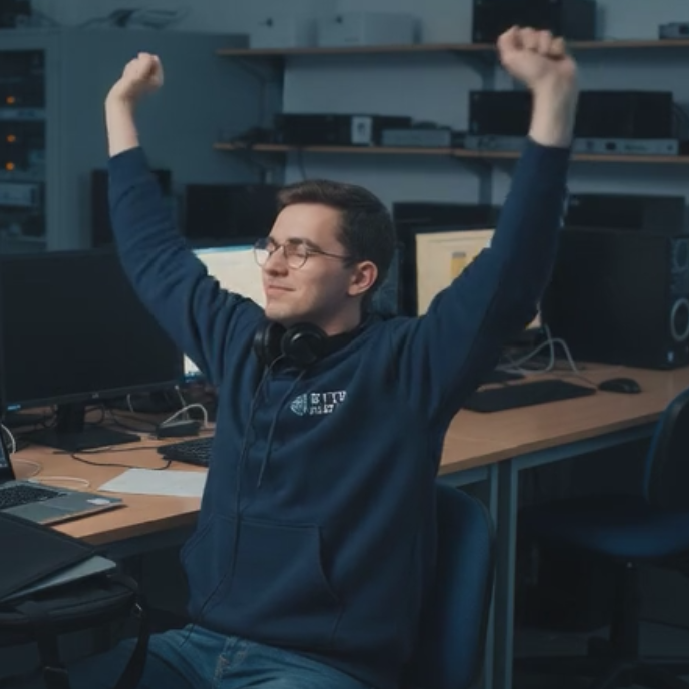}
    }
    \caption{
        \textbf{Case Study 1: GCM for Identity Preservation.} 
        For the prompt "A day of a Computer Science student, in vlog style," the \textbf{Baseline} (top row) exhibits severe identity drift, generating four different people. 
        Our \textbf{CoAgent} (bottom row), empowered by the GCM, maintains a perfectly consistent identity across all shots, visually demonstrating the quantitative gains in \emph{Subject Consistency} from Table~\ref{tab:ablation_study}.
    }
    \label{fig:qual_gcm_consistency}
\end{figure*}
Quantitative metrics (Sec.~\ref{sec:sota_comparison} \& \ref{sec:ablation}) demonstrate that our framework is effective. In this section, we provide qualitative case studies to show \emph{how} and \emph{why} it works.We provide \textbf{extensive} additional qualitative results in the \textbf{Supplementary Material (Sec. C)}, including more comparisons and failure cases.

\begin{figure*}[t!]
    \centering
    
    \begin{tabular}{ p{0.3\textwidth} @{\hspace{3mm}} p{0.3\textwidth} @{\hspace{3mm}} p{0.3\textwidth} }
        
        \centering
        \includegraphics[width=0.80\linewidth]{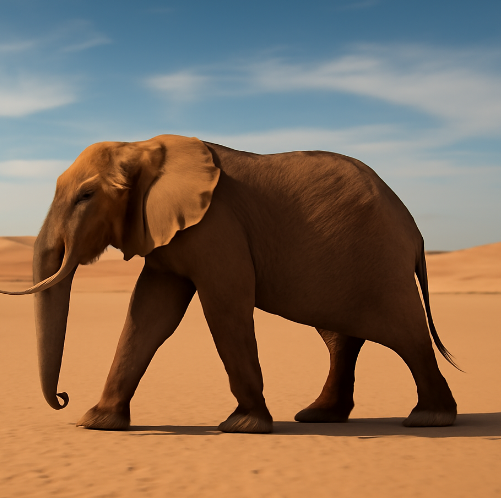}
        \vspace{0.5mm} 
        \includegraphics[width=0.80\linewidth]{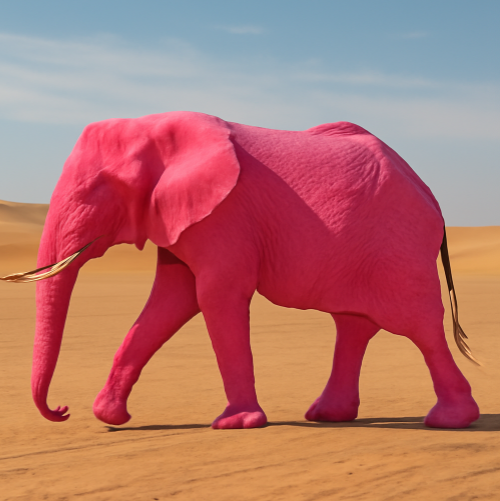}
        \parbox[t]{\linewidth}{\centering \small (a) \textbf{Semantic Correction:} The baseline (top) defaults to common sense (elephant is gray), while our model (bottom) enforces the anti-common sense prompt (elephant is pink).}
        
        & 
        
        \centering
        \includegraphics[width=0.80\linewidth]{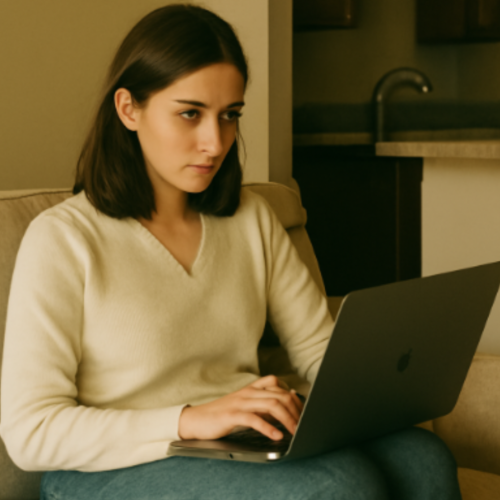}
        \vspace{0.5mm}
        \includegraphics[width=0.80\linewidth]{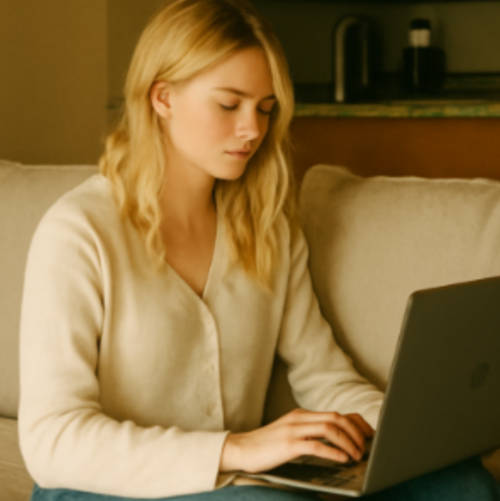}
        \parbox[t]{\linewidth}{\centering \small (b) \textbf{Attribute Correction:} Baseline (top) generates "brown" hair despite the "blonde" prompt. Ours (bottom) corrects this factual error.}

        & 
        
        \centering
        \includegraphics[width=0.80\linewidth]{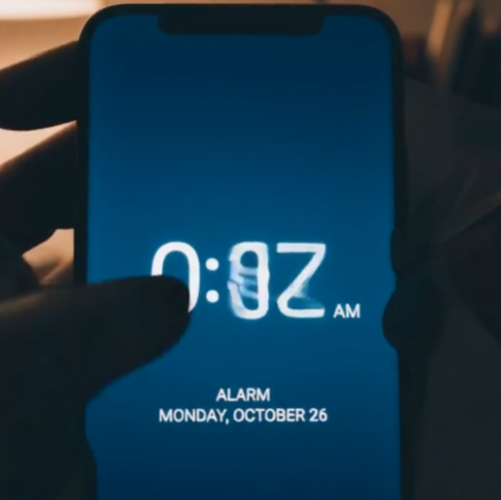}
        \vspace{0.5mm}
        \includegraphics[width=0.80\linewidth]{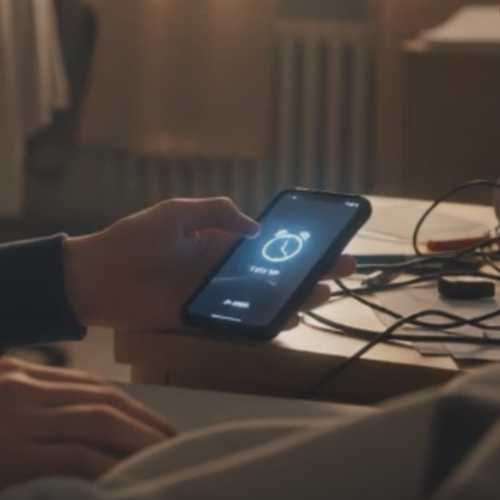}
        \parbox[t]{\linewidth}{\centering \small (c) \textbf{Artifact Correction:} Baseline (top) produces a flickering, illegible clock display. Ours (bottom) regenerates a stable, clear shot.}

    \end{tabular}
    \caption{
        \textbf{Case Study 2: Verifier-driven Multi-Modal Correction.} 
        Our Verifier robustly corrects diverse errors identified in open-loop generation (top row of each pair).
        \textbf{(a) Semantic Correction:} The baseline defaults to common sense, while our Verifier (bottom) enforces the anti-common sense prompt.
        \textbf{(b) Attribute Correction:} The baseline ignores "blonde" while Ours (bottom) corrects the factual error. 
        \textbf{(c) Artifact Correction:} The baseline produces temporal artifacts, while Ours (bottom) regenerates a stable shot.
    }
    \label{fig:qual_verifier_multi}
\end{figure*}

\textbf{Case Study 1: GCM for Identity Preservation.}
To validate the GCM's role in maintaining long-range consistency, we use the complex narrative prompt: "A day of a Computer Science student, in vlog style." As shown in Figure~\ref{fig:qual_gcm_consistency}, the open-loop baseline (which lacks the GCM) fails completely, generating four different individuals across four shots. This visual failure corresponds to its low quantitative score in \emph{Subject Consistency}. In stark contrast, \textbf{CoAgent (Ours)}, empowered by the GCM's explicit entity memory, maintains a perfectly consistent identity for the \emph{same student} throughout the entire multi-shot narrative.

\textbf{Case Study 2: Verifier-driven Multi-Modal Correction.}
To demonstrate the versatility of our \emph{Verifier Agent} ($\mathcal{A}_{verify}$), we present three distinct correction scenarios in Figure~\ref{fig:qual_verifier_multi}.
\begin{itemize}
    \item \textbf{(a) Semantic \& Logic Correction:} Given a prompt with anti-common sense logic ("A pink elephant walks through the desert..."), the baseline synthesizer defaults to common sense, showing a normal gray elephant \textbf{(Fig.~\ref{fig:qual_verifier_multi}(a), top)}. Our Verifier detects this direct contradiction of the prompt's narrative instruction and triggers a correction, resulting in the semantically-correct (though common sense-defying) video of the elephant turning pink \textbf{(Fig.~\ref{fig:qual_verifier_multi}(a), bottom)}. This visually explains the large gains in \emph{Text-Video Alignment}.
    
    \item \textbf{(b) Attribute Correction:} Given "A blonde girl...", the baseline synthesizer incorrectly generates a girl with \emph{brown} hair due to model bias \textbf{(Fig.~\ref{fig:qual_verifier_multi}(b), top)}. The Verifier identifies this factual attribute error (`blonde` vs. `brown`), flags it, and forces a regeneration that correctly depicts the \emph{blonde} girl \textbf{(Fig.~\ref{fig:qual_verifier_multi}(b), bottom)}.
    
    \item \textbf{(c) Temporal Artifact Correction:} Given a prompt involving a phone screen ("...shuts off his phone alarm..."), the baseline generation exhibits severe \emph{Temporal Flickering}, where the time display is corrupted and illegible \textbf{(Fig.~\ref{fig:qual_verifier_multi}(c), top)}. The Verifier detects this high-frequency artifact and regenerates a stable, crisp display \textbf{(Fig.~\ref{fig:qual_verifier_multi}(c), bottom)}. This directly corresponds to the +4.5 point improvement in \emph{Temporal Flickering} shown in Table~\ref{tab:ablation_study}.
\end{itemize}
These cases prove that our Verifier is not a single-purpose tool, but a comprehensive agent that robustly corrects semantic, factual, and temporal errors.



\section{Conclusion}
\label{sec:conclusion}

In this work, we presented \textbf{CoAgent}, a novel collaborative and closed-loop framework that transforms multi-shot video synthesis from a brittle, stateless process into a robust, stateful, and feedback-driven system. We successfully bridged the "narrative gap" by unifying creative planning ($\mathcal{A}_{plan}$), explicit entity-level memory ($\mathcal{M}_{GCM}$), and automated quality control ($\mathcal{A}_{verify}$).

Our primary contributions lie in the architectural shift from open-loop execution to a self-correcting paradigm. The introduction of the Global Context Manager (GCM) proved critical, providing the necessary statefulness to minimize identity drift and scene discontinuity across long sequences. Furthermore, the Verifier Agent's closed-loop mechanism demonstrated its power in rectifying semantic inconsistencies and improving temporal stability via selective regeneration, a capability that open-loop systems fundamentally lack. Quantitatively, CoAgent significantly outperformed state-of-the-art agent-based systems (VISTA) and foundation models (Sora2), establishing new benchmarks for subject consistency, background coherence, and text-video alignment.

Looking ahead, we recognize that true cinematic immersion requires multi-modal coherence. A vital future direction is the integration of \textbf{audio generation and synchronization} into the CoAgent framework. We plan to develop specialized Audio Agents that can generate rhythm-aware soundtracks, sound effects, and dialogue, followed by a multi-modal Verifier to ensure contextual and temporal alignment between the visual and auditory components. This extension will further solidify CoAgent as a holistic platform for high-fidelity, controllable, and fully immersive narrative creation.

\newpage
{
    \small
    \bibliographystyle{ieeenat_fullname}
    \bibliography{main}

@article{wan2025wan,
  title={Wan: Open and advanced large-scale video generative models},
  author={Wan, Team and Wang, Ang and Ai, Baole and Wen, Bin and Mao, Chaojie and Xie, Chen-Wei and Chen, Di and Yu, Feiwu and Zhao, Haiming and Yang, Jianxiao and others},
  journal={arXiv preprint arXiv:2503.20314},
  year={2025}
}

@article{he2024kubrick,
  title={Kubrick: Multimodal agent collaborations for synthetic video generation},
  author={He, Liu and Song, Yizhi and Huang, Hejun and Liu, Pinxin and Tang, Yunlong and Aliaga, Daniel and Zhou, Xin},
  journal={arXiv preprint arXiv:2408.10453},
  year={2024}
}

@inproceedings{wang2025cinemaster,
  title={Cinemaster: A 3d-aware and controllable framework for cinematic text-to-video generation},
  author={Wang, Qinghe and Luo, Yawen and Shi, Xiaoyu and Jia, Xu and Lu, Huchuan and Xue, Tianfan and Wang, Xintao and Wan, Pengfei and Zhang, Di and Gai, Kun},
  booktitle={Proceedings of the Special Interest Group on Computer Graphics and Interactive Techniques Conference Conference Papers},
  pages={1--10},
  year={2025}
}

@inproceedings{yuan2025identity,
  title={Identity-preserving text-to-video generation by frequency decomposition},
  author={Yuan, Shenghai and Huang, Jinfa and He, Xianyi and Ge, Yunyang and Shi, Yujun and Chen, Liuhan and Luo, Jiebo and Yuan, Li},
  booktitle={Proceedings of the Computer Vision and Pattern Recognition Conference},
  pages={12978--12988},
  year={2025}
}

@inproceedings{he-etal-2024-videoscore,
    title = "{V}ideo{S}core: Building Automatic Metrics to Simulate Fine-grained Human Feedback for Video Generation",
    author = "He, Xuan  and
      Jiang, Dongfu  and
      Zhang, Ge  and
      Ku, Max  and
      Soni, Achint  and
      Siu, Sherman  and
      Chen, Haonan  and
      Chandra, Abhranil  and
      Jiang, Ziyan  and
      Arulraj, Aaran  and
      Wang, Kai  and
      Do, Quy Duc  and
      Ni, Yuansheng  and
      Lyu, Bohan  and
      Narsupalli, Yaswanth  and
      Fan, Rongqi  and
      Lyu, Zhiheng  and
      Lin, Bill Yuchen  and
      Chen, Wenhu",
    editor = "Al-Onaizan, Yaser  and
      Bansal, Mohit  and
      Chen, Yun-Nung",
    booktitle = "Proceedings of the 2024 Conference on Empirical Methods in Natural Language Processing",
    month = nov,
    year = "2024",
    address = "Miami, Florida, USA",
    publisher = "Association for Computational Linguistics",
    url = "https://aclanthology.org/2024.emnlp-main.127/",
    doi = "10.18653/v1/2024.emnlp-main.127",
    pages = "2105--2123"
}

@article{zheng2025vbench,
  title={Vbench-2.0: Advancing video generation benchmark suite for intrinsic faithfulness},
      author={Dian Zheng and Ziqi Huang and Hongbo Liu and Kai Zou and Yinan He and Fan Zhang and Yuanhan Zhang and Jingwen He and Wei-Shi Zheng and Yu Qiao and Ziwei Liu},
  journal={arXiv preprint arXiv:2503.21755},
  year={2025},
url="https://arxiv.org/pdf/2503.21755"
}

@article{long2025vista,
  title={VISTA: A Test-Time Self-Improving Video Generation Agent},
  author={Long, Do Xuan and Wan, Xingchen and Nakhost, Hootan and Lee, Chen-Yu and Pfister, Tomas and Ar{\i}k, Sercan {\"O}},
  journal={arXiv preprint arXiv:2510.15831},
  year={2025}
}

@article{cheng2025vpo,
  title={VPO: Aligning Text-to-Video Generation Models with Prompt Optimization},
  author={Cheng, Jiale and Lyu, Ruiliang and Gu, Xiaotao and Liu, Xiao and Xu, Jiazheng and Lu, Yida and Teng, Jiayan and Yang, Zhuoyi and Dong, Yuxiao and Tang, Jie and others},
  journal={arXiv preprint arXiv:2503.20491},
  year={2025},
url="https://arxiv.org/pdf/2503.20491"
}

@article{yang2024cogvideox,
  title={Cogvideox: Text-to-video diffusion models with an expert transformer},
  author={Yang, Zhuoyi and Teng, Jiayan and Zheng, Wendi and Ding, Ming and Huang, Shiyu and Xu, Jiazheng and Yang, Yuanming and Hong, Wenyi and Zhang, Xiaohan and Feng, Guanyu and others},
  journal={arXiv preprint arXiv:2408.06072},
  year={2024}
}

@article{peng2025open,
  title={Open-sora 2.0: Training a commercial-level video generation model in $200 k$},
  author={Peng, Xiangyu and Zheng, Zangwei and Shen, Chenhui and Young, Tom and Guo, Xinying and Wang, Binluo and Xu, Hang and Liu, Hongxin and Jiang, Mingyan and Li, Wenjun and others},
  journal={arXiv preprint arXiv:2503.09642},
  year={2025}
}

@article{madaan2023self,
  title={Self-refine: Iterative refinement with self-feedback},
  author={Madaan, Aman and Tandon, Niket and Gupta, Prakhar and Hallinan, Skyler and Gao, Luyu and Wiegreffe, Sarah and Alon, Uri and Dziri, Nouha and Prabhumoye, Shrimai and Yang, Yiming and others},
  journal={Advances in Neural Information Processing Systems},
  volume={36},
  pages={46534--46594},
  year={2023},
url="https://openreview.net/pdf?id=S37hOerQLB"
}

@misc{openai_sora2_2025,
  author       = {{OpenAI}},
  title        = {Sora 2 is here},
  howpublished = {\url{https://openai.com/index/sora-2/}},
  year         = {2025},
  month        = {sep},
  day          = {30}
}

@article{huang2024genmac,
  title={Genmac: compositional text-to-video generation with multi-agent collaboration},
  author={Huang, Kaiyi and Huang, Yukun and Ning, Xuefei and Lin, Zinan and Wang, Yu and Liu, Xihui},
  journal={arXiv preprint arXiv:2412.04440},
  year={2024}
}

@article{wei2025hollywood,
  title={Hollywood Town: Long-Video Generation via Cross-Modal Multi-Agent Orchestration},
  author={Wei, Zheng and Li, Mingchen and Zhang, Zeqian and Yuan, Ruibin and Hui, Pan and Qu, Huamin and Evans, James and Agrawala, Maneesh and Rao, Anyi},
  journal={arXiv preprint arXiv:2510.22431},
  year={2025}
}

@article{dinkevich2025story2board,
  title={Story2Board: A Training-Free Approach for Expressive Storyboard Generation},
  author={Dinkevich, David and Levy, Matan and Avrahami, Omri and Samuel, Dvir and Lischinski, Dani},
  journal={arXiv preprint arXiv:2508.09983},
  year={2025}
}

@article{ye2023ip,
  title={Ip-adapter: Text compatible image prompt adapter for text-to-image diffusion models},
  author={Ye, Hu and Zhang, Jun and Liu, Sibo and Han, Xiao and Yang, Wei},
  journal={arXiv preprint arXiv:2308.06721},
  year={2023}
}

@article{wang2024instantid,
  title={Instantid: Zero-shot identity-preserving generation in seconds},
  author={Wang, Qixun and Bai, Xu and Wang, Haofan and Qin, Zekui and Chen, Anthony and Li, Huaxia and Tang, Xu and Hu, Yao},
  journal={arXiv preprint arXiv:2401.07519},
  year={2024}
}

@inproceedings{li2024photomaker,
  title={Photomaker: Customizing realistic human photos via stacked id embedding},
  author={Li, Zhen and Cao, Mingdeng and Wang, Xintao and Qi, Zhongang and Cheng, Ming-Ming and Shan, Ying},
  booktitle={Proceedings of the IEEE/CVF conference on computer vision and pattern recognition},
  pages={8640--8650},
  year={2024}
}

@article{fang2024motioncharacter,
  title={Motioncharacter: Identity-preserving and motion controllable human video generation},
  author={Fang, Haopeng and Qiu, Di and Mao, Binjie and Yan, Pengfei and Tang, He},
  journal={arXiv preprint arXiv:2411.18281},
  year={2024}
}

@article{wu2025video,
  title={Video World Models with Long-term Spatial Memory},
  author={Wu, Tong and Yang, Shuai and Po, Ryan and Xu, Yinghao and Liu, Ziwei and Lin, Dahua and Wetzstein, Gordon},
  journal={arXiv preprint arXiv:2506.05284},
  year={2025}
}

@inproceedings{wu2025corgi,
  title={Corgi: Cached Memory Guided Video Generation},
  author={Wu, Xindi and Singer, Uriel and Lin, Zhaojiang and Madotto, Andrea and Xia, Xide and Xu, Yifan and Crook, Paul and Dong, Xin Luna and Moon, Seungwhan},
  booktitle={2025 IEEE/CVF Winter Conference on Applications of Computer Vision (WACV)},
  pages={4585--4594},
  year={2025},
  organization={IEEE}
}

@article{zheng2024videogen,
  title={VideoGen-of-Thought: Step-by-step generating multi-shot video with minimal manual intervention},
  author={Zheng, Mingzhe and Xu, Yongqi and Huang, Haojian and Ma, Xuran and Liu, Yexin and Shu, Wenjie and Pang, Yatian and Tang, Feilong and Chen, Qifeng and Yang, Harry and others},
  journal={arXiv preprint arXiv:2412.02259},
  year={2024}
}

@inproceedings{wu2024self,
  title={Self-correcting llm-controlled diffusion models},
  author={Wu, Tsung-Han and Lian, Long and Gonzalez, Joseph E and Li, Boyi and Darrell, Trevor},
  booktitle={Proceedings of the IEEE/CVF Conference on Computer Vision and Pattern Recognition},
  pages={6327--6336},
  year={2024}
}

@article{meng2025holocine,
  title={HoloCine: Holistic Generation of Cinematic Multi-Shot Long Video Narratives},
  author={Meng, Yihao and Ouyang, Hao and Yu, Yue and Wang, Qiuyu and Wang, Wen and Cheng, Ka Leong and Wang, Hanlin and Li, Yixuan and Chen, Cheng and Zeng, Yanhong and others},
  journal={arXiv preprint arXiv:2510.20822},
  year={2025}
}

@inproceedings{
z1,
title={{KABB}: Knowledge-Aware Bayesian Bandits for Dynamic Expert Coordination in Multi-Agent Systems},
author={Jusheng Zhang and Zimeng Huang and Yijia Fan and Ningyuan Liu and Mingyan Li and Zhuojie Yang and Jiawei Yao and Jian Wang and Keze Wang},
booktitle={Forty-second International Conference on Machine Learning},
year={2025},
url={https://openreview.net/forum?id=AKvy9a4jho}
}

@inproceedings{
z2,
title={{GAM}-Agent: Game-Theoretic and Uncertainty-Aware Collaboration for Complex Visual Reasoning},
author={Jusheng Zhang and Yijia Fan and Wenjun Lin and Ruiqi Chen and Haoyi Jiang and Wenhao Chai and Jian Wang and Keze Wang},
booktitle={The Thirty-ninth Annual Conference on Neural Information Processing Systems},
year={2025},
url={https://openreview.net/forum?id=EKJhU5ioSo}
}

@inproceedings{Z3,
  title         = {{CF-VLM}: Counterfactual Vision-Language Fine-tuning},
  author        = {Jusheng Zhang and Kaitong Cai and Yijia Fan and Jian Wang and Keze Wang},
  booktitle     = {Advances in Neural Information Processing Systems},
  year          = {2025},
  url           = {https://neurips.cc/virtual/2025/poster/120284},
  eprint        = {2506.17267},
  archivePrefix = {arXiv},
  primaryClass  = {cs.LG},
  doi           = {10.48550/arXiv.2506.17267},
  note          = {OpenReview: https://openreview.net/forum?id=0qGtaRTsCo}
}

@inproceedings{
z4,
title={{MAT}-Agent: Adaptive Multi-Agent Training Optimization},
author={Jusheng Zhang and Kaitong Cai and Yijia Fan and Ningyuan Liu and Keze Wang},
booktitle={The Thirty-ninth Annual Conference on Neural Information Processing Systems},
year={2025},
url={https://openreview.net/forum?id=YDWRTYgR79}
}

@inproceedings{
Z5,
title={Tri-{MARF}: A Tri-Modal Multi-Agent Responsive Framework for Comprehensive 3D Object Annotation},
author={Jusheng Zhang and Yijia Fan and Zimo Wen and Jian Wang and Keze Wang},
booktitle={The Thirty-ninth Annual Conference on Neural Information Processing Systems},
year={2025},
url={https://openreview.net/forum?id=YGIbwfNWot}
}

@misc{z6,
  title         = {MM-CoT:A Benchmark for Probing Visual Chain-of-Thought Reasoning in Multimodal Models},
  author        = {Jusheng Zhang and Kaitong Cai and Xiaoyang Guo and Sidi Liu and Qinhan Lv and Ruiqi Chen and Jing Yang and Yijia Fan and Xiaofei Sun and Jian Wang and Ziliang Chen and Liang Lin and Keze Wang},
  year          = {2025},
  eprint        = {2512.08228},
  archivePrefix = {arXiv},
  primaryClass  = {cs.CV},
  doi           = {10.48550/arXiv.2512.08228},
  note          = {arXiv:2512.08228},
  url           = {https://arxiv.org/abs/2512.08228}
}

@misc{z7,
  title         = {HybridToken-VLM: Hybrid Token Compression for Vision-Language Models},
  author        = {Jusheng Zhang and Xiaoyang Guo and Kaitong Cai and Qinhan Lv and Yijia Fan and Wenhao Chai and Jian Wang and Keze Wang},
  year          = {2025},
  eprint        = {2512.08240},
  archivePrefix = {arXiv},
  primaryClass  = {cs.CV},
  doi           = {10.48550/arXiv.2512.08240},
  note          = {arXiv:2512.08240},
  url           = {https://arxiv.org/abs/2512.08240}
}

@misc{z8,
  title         = {Kolmogorov-Arnold Fourier Networks},
  author        = {Jusheng Zhang and Yijia Fan and Kaitong Cai and Keze Wang},
  year          = {2025},
  eprint        = {2502.06018},
  archivePrefix = {arXiv},
  primaryClass  = {cs.LG},
  doi           = {10.48550/arXiv.2502.06018},
  note          = {arXiv:2502.06018},
  url           = {https://arxiv.org/abs/2502.06018}
}

@misc{z9,
      title={DrDiff: Dynamic Routing Diffusion with Hierarchical Attention for Breaking the Efficiency-Quality Trade-off}, 
      author={Jusheng Zhang and Yijia Fan and Kaitong Cai and Zimeng Huang and Xiaofei Sun and Jian Wang and Chengpei Tang and Keze Wang},
      year={2025},
      eprint={2509.02785},
      archivePrefix={arXiv},
      primaryClass={cs.CL},
      url={https://arxiv.org/abs/2509.02785}, 
}

@misc{z10,
      title={OSC: Cognitive Orchestration through Dynamic Knowledge Alignment in Multi-Agent LLM Collaboration}, 
      author={Jusheng Zhang and Yijia Fan and Kaitong Cai and Xiaofei Sun and Keze Wang},
      year={2025},
      eprint={2509.04876},
      archivePrefix={arXiv},
      primaryClass={cs.AI},
      url={https://arxiv.org/abs/2509.04876}, 
}

@misc{z11,
      title={Learning Dynamics of VLM Finetuning}, 
      author={Jusheng Zhang and Kaitong Cai and Jing Yang and Keze Wang},
      year={2025},
      eprint={2510.11978},
      archivePrefix={arXiv},
      primaryClass={cs.LG},
      url={https://arxiv.org/abs/2510.11978}, 
}

@misc{z12,
      title={Failure-Driven Workflow Refinement}, 
      author={Jusheng Zhang and Kaitong Cai and Qinglin Zeng and Ningyuan Liu and Stephen Fan and Ziliang Chen and Keze Wang},
      year={2025},
      eprint={2510.10035},
      archivePrefix={arXiv},
      primaryClass={cs.AI},
      url={https://arxiv.org/abs/2510.10035}, 
}

@misc{z13,
      title={Top-Down Semantic Refinement for Image Captioning}, 
      author={Jusheng Zhang and Kaitong Cai and Jing Yang and Jian Wang and Chengpei Tang and Keze Wang},
      year={2025},
      eprint={2510.22391},
      archivePrefix={arXiv},
      primaryClass={cs.CV},
      url={https://arxiv.org/abs/2510.22391}, 
}

@misc{z14,
  title        = {LLM-CAS: Dynamic Neuron Perturbation for Real-Time Hallucination Correction},
  author       = {Jensen Zhang and Ningyuan Liu and Yijia Fan and Zihao Huang and Qinglin Zeng and Kaitong Cai and Jian Wang and Keze Wang},
  year         = {2025},
  eprint       = {2512.18623},
  archivePrefix= {arXiv},
  primaryClass = {cs.CL},
  doi          = {10.48550/arXiv.2512.18623},
  note         = {arXiv:2512.18623},
  url          = {https://arxiv.org/abs/2512.18623}
}

@misc{z15,
      title={DepthSSC: Monocular 3D Semantic Scene Completion via Depth-Spatial Alignment and Voxel Adaptation}, 
      author={Jiawei Yao and Jusheng Zhang and Xiaochao Pan and Tong Wu and Canran Xiao},
      year={2024},
      eprint={2311.17084},
      archivePrefix={arXiv},
      primaryClass={cs.CV},
      url={https://arxiv.org/abs/2311.17084}, 
}

@inproceedings{z16,
    title = "{CCG}: Rare-Label Prediction via Neural {SEM}{--}Driven Causal Game",
    author = "Fan, Yijia  and
      Zhang, Jusheng  and
      Cai, Kaitong  and
      Yang, Jing  and
      Wang, Keze",
    editor = "Christodoulopoulos, Christos  and
      Chakraborty, Tanmoy  and
      Rose, Carolyn  and
      Peng, Violet",
    booktitle = "Findings of the Association for Computational Linguistics: EMNLP 2025",
    month = nov,
    year = "2025",
    address = "Suzhou, China",
    publisher = "Association for Computational Linguistics",
    url = "https://aclanthology.org/2025.findings-emnlp.331/",
    doi = "10.18653/v1/2025.findings-emnlp.331",
    pages = "6243--6256",
    ISBN = "979-8-89176-335-7"
}

@misc{z17,
      title={3DAlign-DAER: Dynamic Attention Policy and Efficient Retrieval Strategy for Fine-grained 3D-Text Alignment at Scale}, 
      author={Yijia Fan and Jusheng Zhang and Kaitong Cai and Jing Yang and Jian Wang and Keze Wang},
      year={2025},
      eprint={2511.13211},
      archivePrefix={arXiv},
      primaryClass={cs.CV},
      url={https://arxiv.org/abs/2511.13211}, 
}

@misc{z18,
      title={Cost-Effective Communication: An Auction-based Method for Language Agent Interaction}, 
      author={Yijia Fan and Jusheng Zhang and Kaitong Cai and Jing Yang and Chengpei Tang and Jian Wang and Keze Wang},
      year={2025},
      eprint={2511.13193},
      archivePrefix={arXiv},
      primaryClass={cs.AI},
      url={https://arxiv.org/abs/2511.13193}, 
}

@misc{z19,
      title={RaCoT: Plug-and-Play Contrastive Example Generation Mechanism for Enhanced LLM Reasoning Reliability}, 
      author={Kaitong Cai and Jusheng Zhang and Yijia Fan and Jing Yang and Keze Wang},
      year={2025},
      eprint={2510.22710},
      archivePrefix={arXiv},
      primaryClass={cs.AI},
      url={https://arxiv.org/abs/2510.22710}, 
}

@article{z20,
author = {Li, Xiaohua and Zhang, Jusheng and Safara, Fatemeh},
title = {Improving the Accuracy of Diabetes Diagnosis Applications through a Hybrid Feature Selection Algorithm},
year = {2021},
issue_date = {Feb 2023},
publisher = {Kluwer Academic Publishers},
address = {USA},
volume = {55},
number = {1},
issn = {1370-4621},
url = {https://doi.org/10.1007/s11063-021-10491-0},
doi = {10.1007/s11063-021-10491-0},
journal = {Neural Process. Lett.},
month = mar,
pages = {153–169},
numpages = {17}
}

@misc{z22,
  title         = {FlashVLM: Text-Guided Visual Token Selection for Large Multimodal Models},
  author        = {Kaitong Cai and Jusheng Zhang and Jing Yang and Yijia Fan and Pengtao Xie and Jian Wang and Keze Wang},
  year          = {2025},
  eprint        = {2512.20561},
  archivePrefix = {arXiv},
  primaryClass  = {cs.CV},
  doi           = {10.48550/arXiv.2512.20561},
  note          = {arXiv:2512.20561},
  url           = {https://arxiv.org/abs/2512.20561}
}
}


\clearpage
\setcounter{page}{1}
\maketitlesupplementary

\section{Appendix A. Implementation Details}
\label{sec:imp_details}

As referenced in Sec.~4.1 of the main paper, we provide detailed implementation specifications for CoAgent, including the exact prompts used for agent orchestration, backbone model configurations, and evaluation setups.

\subsection{Agent Prompts and Instructions}
\label{subsec:prompts}

To facilitate reproducibility, we present the core system instructions used for the Storyboard Planner ($\mathcal{A}_{plan}$), Global Context Manager (GCM), and Verifier Agent ($\mathcal{A}_{verify}$). Note that for experiments on Chinese prompts, equivalent translated instructions were utilized.

\subsubsection{Storyboard Planner}
The Storyboard Planner (powered by Gemini-2.5-Flash) is the architectural brain of CoAgent. It decomposes abstract concepts into filmable shots with explicit mode selection (\texttt{ff2v} vs. \texttt{flf2v}).

\begin{promptbox}[title=System Instruction: Storyboard Planner]
    \textbf{Role:} You are a professional screenwriter and storyboard planner.
    
    \textbf{Task:} Infer a reasonable number of shots $N$ from pacing and target duration, then output EXACTLY $N$ detailed shots with filmable visuals. For each shot, decide the video generation mode: \texttt{ff2v} (only first frame) or \texttt{flf2v} (first \& last frames).
    
    \textbf{Guidelines:}
    \begin{itemize}
        \item Language must follow the user's language.
        \item EXACTLY $N$ shots; choose $N$ based on pacing and duration.
        \item Filmable and concrete; avoid abstract emotions.
        \item Pick \texttt{flf2v} when the shot requires a clear start$\rightarrow$end transition or long continuous motion; otherwise \texttt{ff2v}.
        \item If generation mode is \texttt{flf2v}, you MUST provide \texttt{last\_frame\_prompt} that logically matches motion and can connect to the next shot.
        \item If generation mode is \texttt{ff2v} and seamless continuity to next is needed, set \texttt{connect\_to\_next=true}.
        \item Output STRICT JSON only.
    \end{itemize}
\end{promptbox}

The strictly enforced JSON output schema is detailed below:

\begin{promptbox}[title=Output Data Structure Requirement]
    Instead of raw text, the Planner is instructed to output a strict JSON object containing the following key components:
    
    \begin{itemize}
        \item \textbf{Global Metadata:} Title, Target Audience, Genre, Style, Pacing, and Logline.
        \item \textbf{Character Registry:} A list of characters, each defined by a unique \texttt{id}, \texttt{name}, and \texttt{static\_features} (to ensure appearance consistency across shots).
        \item \textbf{Shot Sequence:} An ordered list of shots. Each shot object includes:
        \begin{itemize}
            \item \textbf{Visual Plan:}
            \begin{itemize}
                \item \texttt{generation\_mode}: Selection between \texttt{ff2v} (standard) or \texttt{flf2v} (transition-heavy).
                \item \texttt{camera\_angle} \& \texttt{lighting}: Cinematic directives.
                \item \texttt{first/last\_frame\_prompt}: Detailed visual descriptions for keyframes.
            \end{itemize}
            \item \textbf{Transition Logic:} A \texttt{connect\_to\_next} boolean flag to enforce temporal continuity.
            \item \textbf{Audio:} Sound effects or dialogue description.
        \end{itemize}
    \end{itemize}
\end{promptbox}

\subsubsection{Global Context Manager (GCM)}
The GCM generates high-fidelity reference portraits. These "Master Portraits" serve as the visual anchor ($\mathcal{M}_{GCM}$) for all subsequent consistency checks.

\begin{promptbox}[title=Image Generation Prompt: GCM]
    ``Generate a clean character portrait (setup shot):
    \begin{itemize}
        \item \textbf{Character:} [Character Description]
        \item \textbf{Style:} [Style Description or 'Default']
        \item \textbf{Requirements:} Front view, pure white background, cinematic lighting, clear facial features, full body or half body shot.''
    \end{itemize}
\end{promptbox}

\subsubsection{Verifier Agent}
The Verifier ($\mathcal{A}_{verify}$) executes a two-stage audit for every generated shot.

\textbf{Stage 1: Semantic Consistency.}
\begin{promptbox}[title=Verifier Instruction: Semantic Check]
    ``Act as a professional film reviewer. Judge solely based on the provided frame:
    \\ \textbf{Script Description:} [Script Description]
    \\ \textbf{Requirement:} Answer only PASS or FAIL; if FAIL, briefly explain the mismatch point (under 20 words).''
\end{promptbox}

\textbf{Stage 2: Visual Identity Consistency.}
\begin{promptbox}[title=Verifier Instruction: Identity Check]
    ``Strictly compare whether the same character in the two images is consistent:
    \\ \textbf{Image 1:} Master Portrait (GCM Reference); 
    \\ \textbf{Image 2:} Video Frame.
    \\ \textbf{Requirement:} Answer only PASS or FAIL; if FAIL, briefly explain the difference (under 20 words).''
\end{promptbox}

\subsection{Model Configurations and Hardware}
\label{subsec:hyperparams}

Our Synthesis Module ($\mathcal{A}_{synth}$) utilizes the Wan2.1 family. Depending on the Planner's decision, we switch between three distinct configurations as shown in Table \ref{tab:wan_params}.

\begin{table}[htbp]
    \centering
    \small
    \caption{\textbf{Hyperparameters for Wan2.1 Synthesis Modes.} We dynamically switch checkpoints based on the narrative requirement.}
    \label{tab:wan_params}
    \resizebox{\columnwidth}{!}{
        \begin{tabular}{l|cccc}
            \toprule
            \textbf{Mode} & \textbf{Checkpoint Variant} & \textbf{Resolution} & \textbf{Steps} & \textbf{Conditioning} \\
            \midrule
            \textbf{T2V} & Wan2.1-T2V-14B & $832 \times 480$ & 40 & Text Prompt \\
            \textbf{FF2V} & Wan2.1-I2V-14B-480P & $832 \times 480$ & 30 & Text + First Frame \\
            \textbf{FLF2V} & Wan2.1-FLF2V-14B-720P & $1280 \times 720$ & 30 & Text + First/Last Frames \\
            \bottomrule
        \end{tabular}
    }
\end{table}

\paragraph{Hardware and Efficiency.}
All experiments were conducted on a local workstation equipped with a single \textbf{NVIDIA RTX 6000 (Blackwell Architecture, 96GB VRAM)}. Under this configuration, the average inference time for generating a single coherent shot (including agent reasoning and model loading) is approximately \textbf{8 minutes and 14 seconds}.

\subsection{Evaluation Setup: Sora2 Comparison and Metric Bias Correction}
\label{subsec:sora_watermark}

To ensure a fair quantitative comparison with the closed-source Sora2 model (referenced in Table 2 of the main paper), we investigated the impact of platform-specific watermarks on automated evaluation metrics (specifically VideoScore [5]).

\paragraph{Discovery of Watermark Bias.}
During our preliminary evaluation, we observed an anomaly where Sora2 samples containing the official dynamic watermark achieved suspiciously high scores (e.g., Visual Quality $>3.7$). We hypothesized that the evaluation model might have learned a spurious correlation between the specific watermark pattern and "high quality" labels during its training, effectively acting as a "metric hack."

\paragraph{Validation Experiment.}
To verify this hypothesis and justify our preprocessing strategy, we conducted a controlled "Remove-and-Restore" experiment on a subset of Sora2 generated videos ($N=50$). We compared three conditions:
\begin{enumerate}
    \item \textbf{Original:} The raw video with the platform watermark.
    \item \textbf{Inpainted (Used in Paper):} The watermark was removed using a video inpainting model.
    \item \textbf{Re-watermarked (Control):} We took the \textit{Inpainted} video from step 2 and overlaid the watermark again.
\end{enumerate}

The results are presented in Table \ref{tab:watermark_bias}. Removing the watermark caused the \textit{Visual Quality} score to drop significantly from 3.781 to 2.734. Crucially, when the watermark was re-applied to the \textit{same inpainted pixels} (Control), the score rebounded to 3.766, which is almost identical to the original. This proves that the drop in quality is not due to the inpainting damage, but due to the removal of the "watermark bias."

\begin{table}[htbp]
    \centering
    \small
    \caption{\textbf{Impact of Watermark on Evaluation Metrics.} We compare the Original Sora2 videos, the Re-watermarked control group, and the Inpainted version. The scores show a "U-shaped" pattern: removing the watermark drops the score significantly, while adding it back restores the high score. This confirms the metric bias. We use the \textbf{Inpainted (Unbiased)} version for the main paper comparison.}
    \label{tab:watermark_bias}
    \resizebox{\columnwidth}{!}{
        \begin{tabular}{l|cc|c}
            \toprule
            & \multicolumn{2}{c|}{\textbf{Biased (w/ Watermark)}} & \textbf{Unbiased (No Watermark)} \\
            \textbf{Metric} & \textbf{1. Original} & \textbf{3. Control} & \textbf{2. Inpainted (Ours)} \\
            \midrule
            Visual Quality & 3.781 & 3.766 & \textbf{2.734} \\
            Temporal Consistency & 3.578 & 3.531 & \textbf{2.484} \\
            Dynamic Degree & 3.766 & 3.766 & \textbf{2.984} \\
            Text-Video Alignment & 2.969 & 3.031 & \textbf{2.469} \\
            Factual Consistency & 3.484 & 3.469 & \textbf{2.172} \\
            \bottomrule
        \end{tabular}
    }
\end{table}

\paragraph{Conclusion.}
The experiment confirms that the watermark artificially inflates metrics by approximately $+1.0$ point. Comparing our method against the raw Sora2 would be scientifically invalid. Therefore, we adopt the \textbf{Inpainted (Unbiased)} setting as the ground truth for the Sora2 baseline reported in the main paper.

\section{Appendix B. Additional Ablation Studies}
\label{sec:add_ablation}

As promised in Sec.~4.3 of the main paper, we provide further investigations into the generalization capability of our framework and a deeper component analysis.

\subsection{Backbone Generalization}
\label{subsec:backbone_gen}

A key design goal of CoAgent is model-agnosticism. To validate this, we integrated our framework with two other open-source video generation backbones: \textbf{CogVideoX-5B} and the recently released \textbf{LongCat-Video}.

Table \ref{tab:backbone_gen} demonstrates the quantitative results on VBench. CoAgent consistently improves the \textit{Subject Consistency} and \textit{Motion Smoothness} across all tested backbones. Notably, for \textbf{LongCat-Video}, incorporating the CoAgent workflow yields a performance boost of +4.2\% in subject consistency, effectively bridging the gap between efficient open-source models and proprietary SOTA performance.

\begin{table}[htbp]
    \centering
    \small
    \caption{\textbf{Backbone Generalization.} Performance gains of CoAgent applied to different foundation models (VBench scores). We observe consistent improvements across architectures, with LongCat-Video showing significant gains in consistency when governed by our GCM.}
    \label{tab:backbone_gen}
    \resizebox{\columnwidth}{!}{
        \begin{tabular}{l|ccc}
            \toprule
            \textbf{Model / Configuration} & \textbf{Subj. Cons.} & \textbf{Bg. Cons.} & \textbf{Motion Smooth.} \\
            \midrule
            \textit{Backbone 1: CogVideoX-5B} & & & \\
            \quad Baseline (Original) & 89.10 & 91.30 & 94.20 \\
            \quad \textbf{+ CoAgent (Ours)} & \textbf{93.80} & \textbf{94.40} & \textbf{97.00} \\
            \midrule
            \textit{Backbone 2: Wan2.1 (Main)} & & & \\
            \quad Baseline (Original) & 90.60 & 95.50 & 94.80 \\
            \quad \textbf{+ CoAgent (Ours)} & \textbf{94.70} & \textbf{96.50} & \textbf{99.40} \\
            \midrule
            \textit{Backbone 3: LongCat-Video} & & & \\
            \quad Baseline (Original) & 89.85 & 93.40 & 95.50 \\
            \quad \textbf{+ CoAgent (Ours)} & \textbf{94.20} & \textbf{95.90} & \textbf{98.90} \\
            \bottomrule
        \end{tabular}
    }
\end{table}

\subsection{Component Analysis: FLF2V vs. FF2V}
\label{subsec:ablation_mode}

The \texttt{flf2v} (First-and-Last-Frame to Video) mode is a critical innovation for maintaining narrative pacing. In this mode, the Planner hallucinates the last frame of a shot to serve as a "goal anchor," ensuring the video transitions smoothly to the next scene.

To quantify its impact, we compared the full CoAgent system against a variant where the Planner is restricted to use only \texttt{ff2v} (First-Frame only conditioning). As shown in Table \ref{tab:flf2v_ablation}, removing the bi-directional constraint leads to a sharp drop in \textit{Motion Smoothness} (-3.2\%) and \textit{Temporal Style}, as the model lacks guidance on "where to end," often resulting in abrupt cuts or wandering narratives.

\begin{table}[htbp]
    \centering
    \small
    \caption{\textbf{Impact of FLF2V Mode.} We disable the bi-directional conditioning (FLF2V) and force the model to use only FF2V. The drop in metrics indicates that knowing the "ending" is crucial for smooth transitions.}
    \label{tab:flf2v_ablation}
    \resizebox{\columnwidth}{!}{
        \begin{tabular}{l|ccc}
            \toprule
            \textbf{Configuration} & \textbf{Motion Smooth.} & \textbf{Temp. Style} & \textbf{Subj. Cons.} \\
            \midrule
            CoAgent (w/o FLF2V) & 96.20 & 94.10 & 93.80 \\
            \textbf{CoAgent (Full)} & \textbf{99.40} & \textbf{98.50} & \textbf{94.70} \\
            \bottomrule
        \end{tabular}
    }
\end{table}

\subsection{Regeneration Efficiency Analysis}
\label{subsec:efficiency}

A common concern with closed-loop systems is the computational cost of regeneration. We analyzed the number of iterations required for a shot to pass the Verifier. 
\begin{itemize}
    \item \textbf{Pass Rate:} On average, \textbf{72\%} of shots pass on the first attempt ($N=1$).
    \item \textbf{Correction Cost:} For the remaining 28\%, the average number of additional regeneration turns is only \textbf{1.4}. 
    \item \textbf{Convergence:} We set a hard limit of $N=3$. In our experiments, 98\% of shots converge to a satisfactory quality within this limit, demonstrating that the Verifier provides effective, actionable feedback rather than engaging in random search.
\end{itemize}

\section{Appendix C. Additional Qualitative Results}
\label{sec:qual_results}

In this section, we present extended visual comparisons and an analysis of system limitations. We focus on long-horizon consistency compared against the proprietary state-of-the-art model, Sora2, and discuss failure cases related to physical dynamics.

\subsection{Comparison with Sora2}
\label{subsec:sora_comparison}

\begin{figure*}[htbp]
    \centering
    \begin{minipage}{\textwidth}
        \centering
        \textbf{Model A: Sora2 (Baseline)} \\
        \vspace{2pt}
         \includegraphics[width=0.19\textwidth]{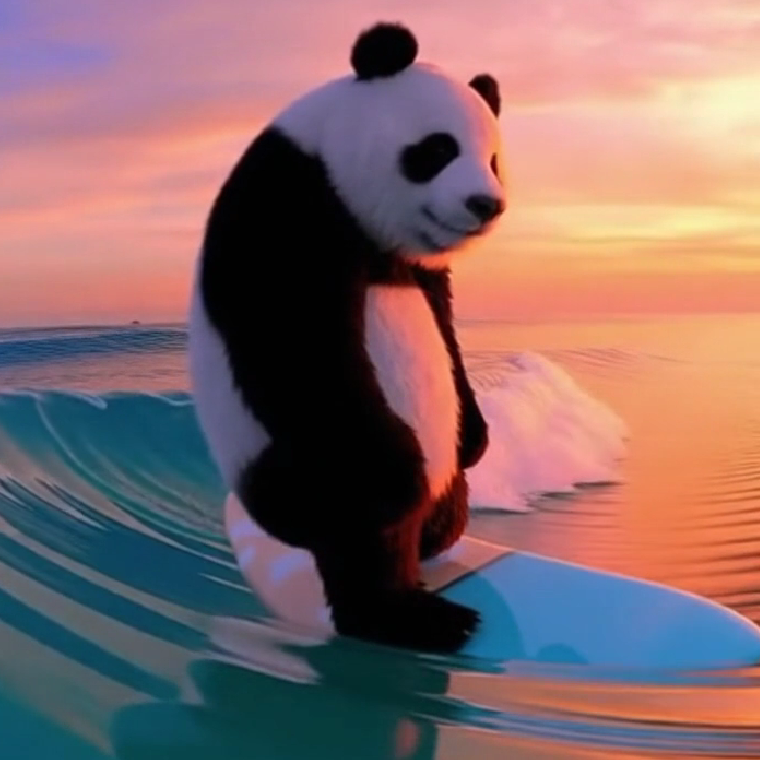} \hfill
         \includegraphics[width=0.19\textwidth]{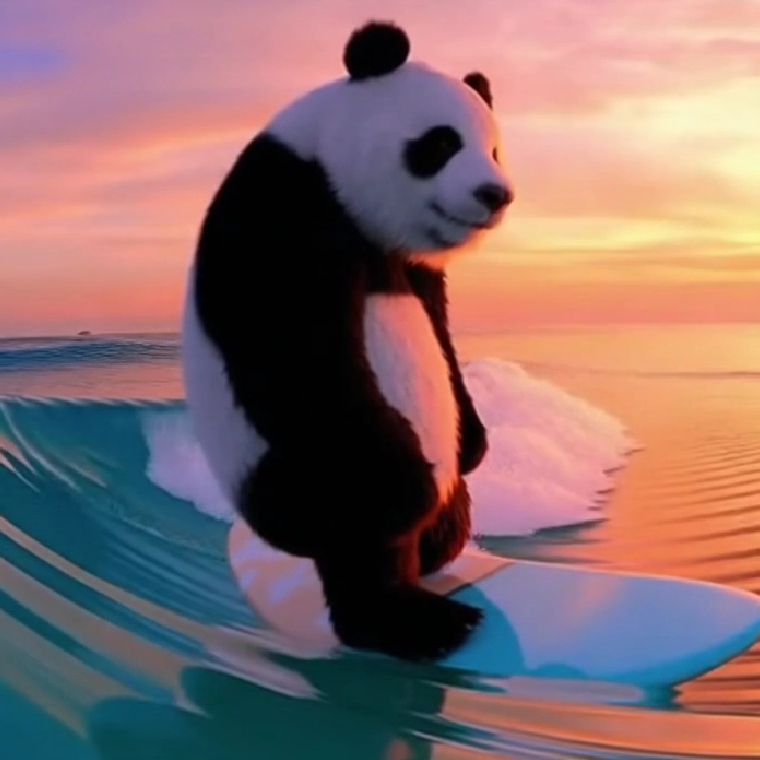} \hfill
        \includegraphics[width=0.19\textwidth]{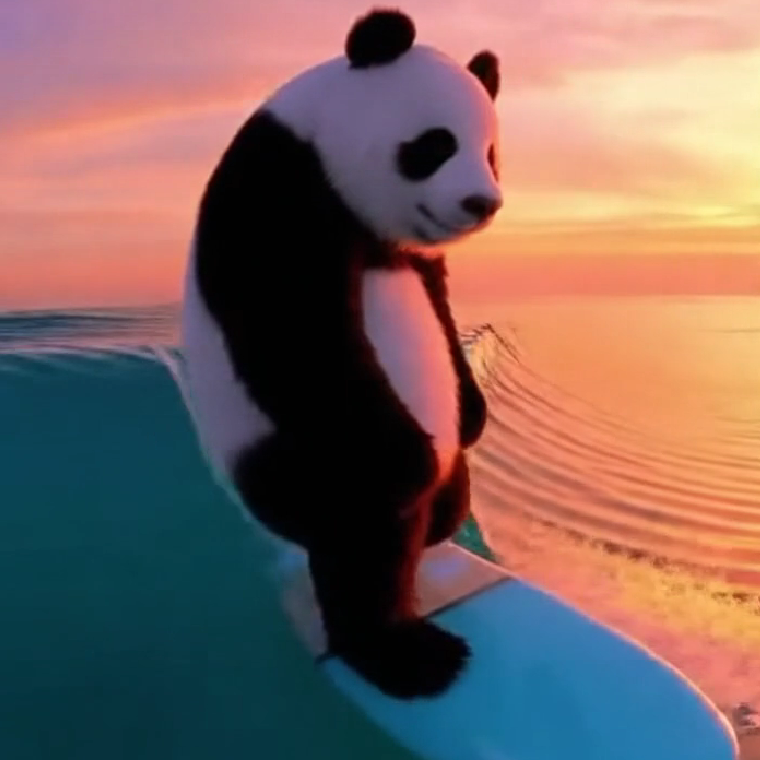} \hfill
        \includegraphics[width=0.19\textwidth]{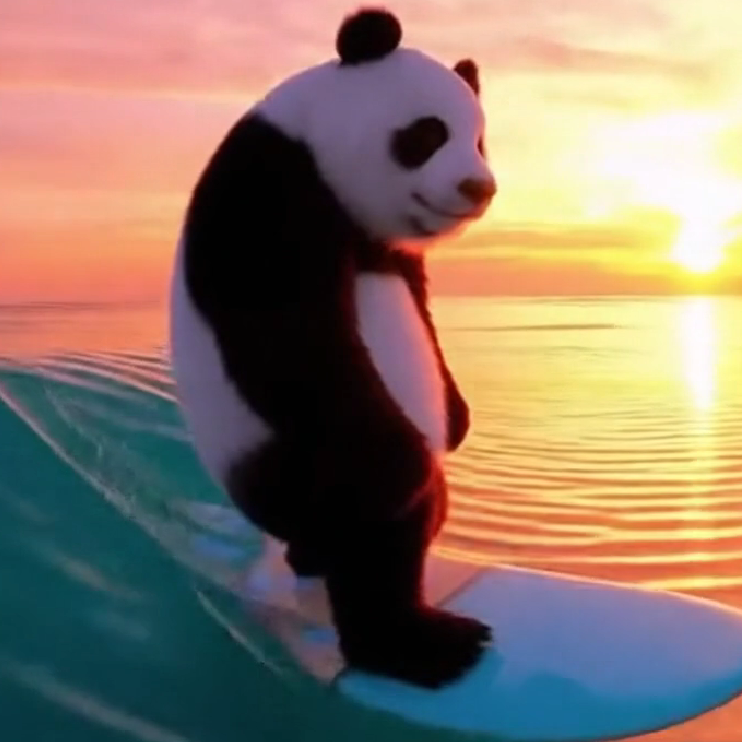} \hfill
        \includegraphics[width=0.19\textwidth]{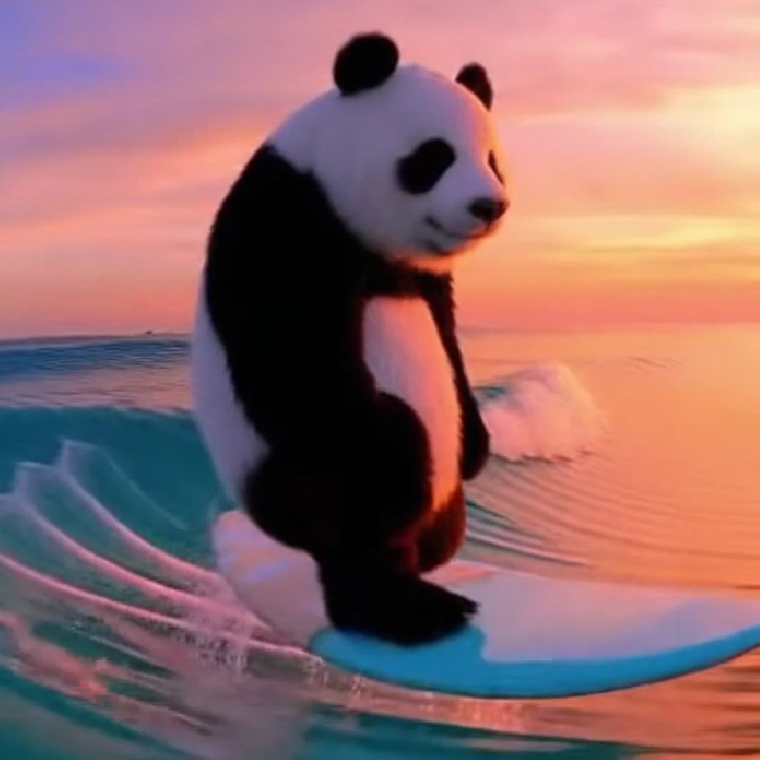}
    \end{minipage}
    
    \vspace{0.5em} 
    
    \begin{minipage}{\textwidth}
        \centering
        \textbf{Model B: CoAgent (Ours)} \\
        \vspace{2pt}
        \includegraphics[width=0.19\textwidth]{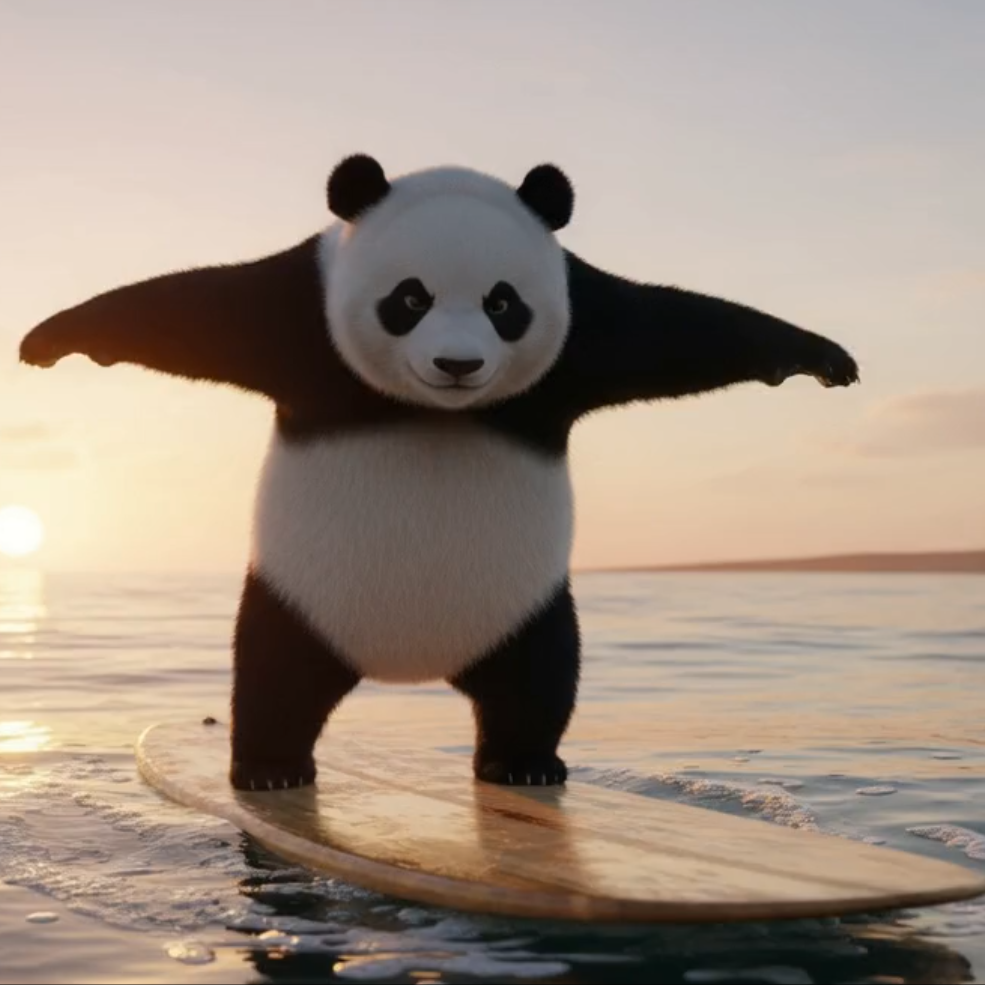} \hfill
        \includegraphics[width=0.19\textwidth]{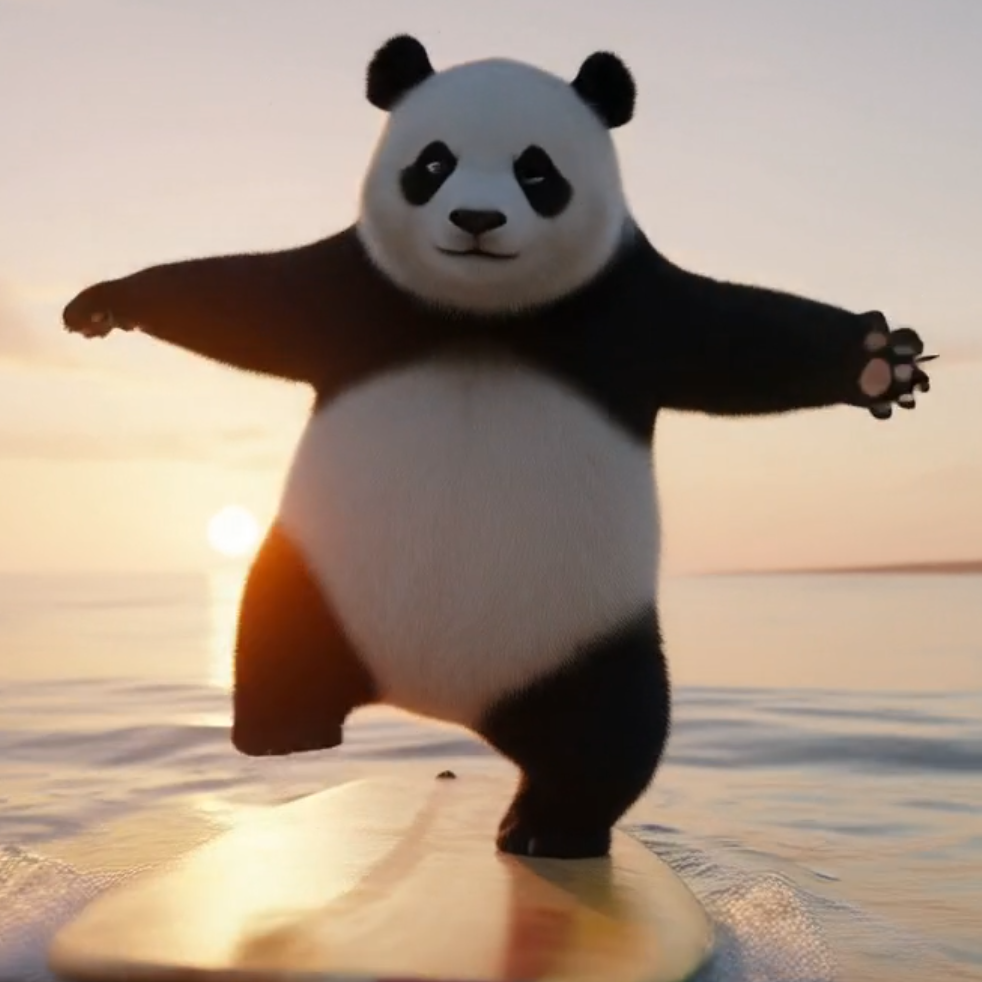} \hfill
        \includegraphics[width=0.19\textwidth]{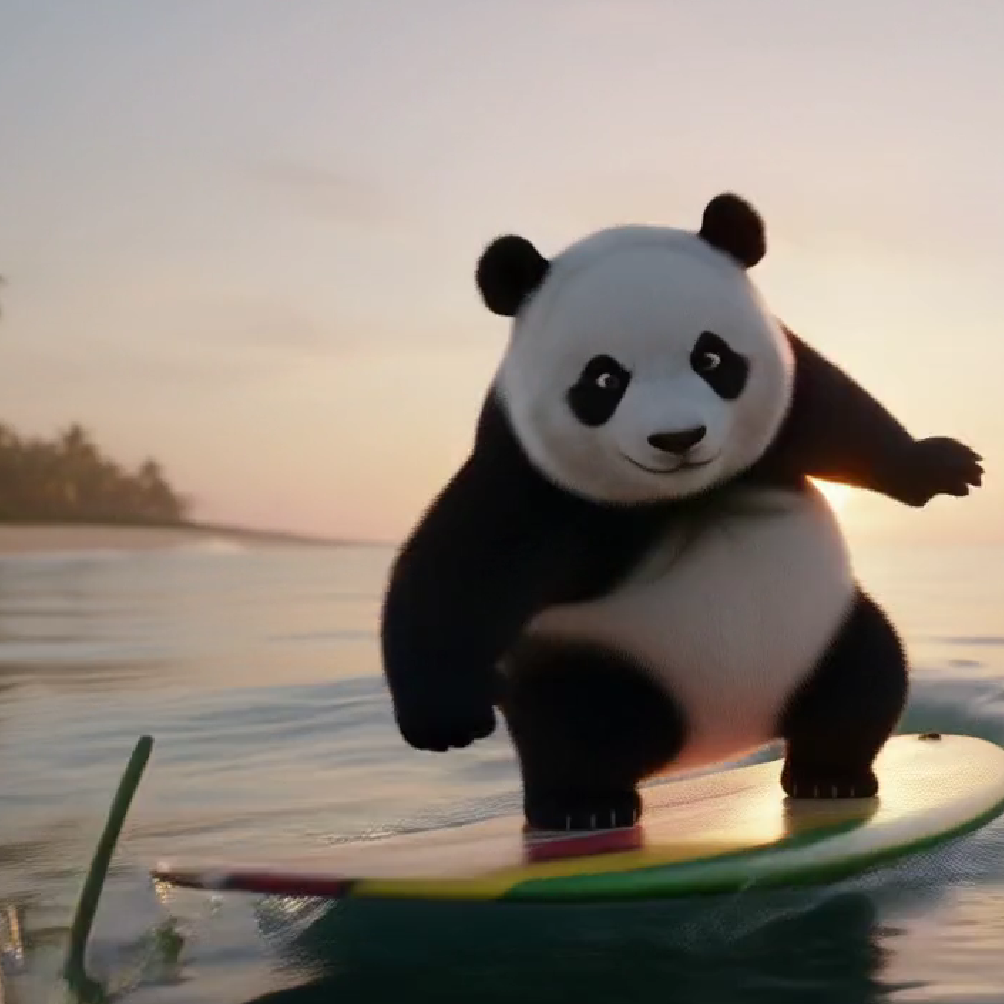} \hfill
        \includegraphics[width=0.19\textwidth]{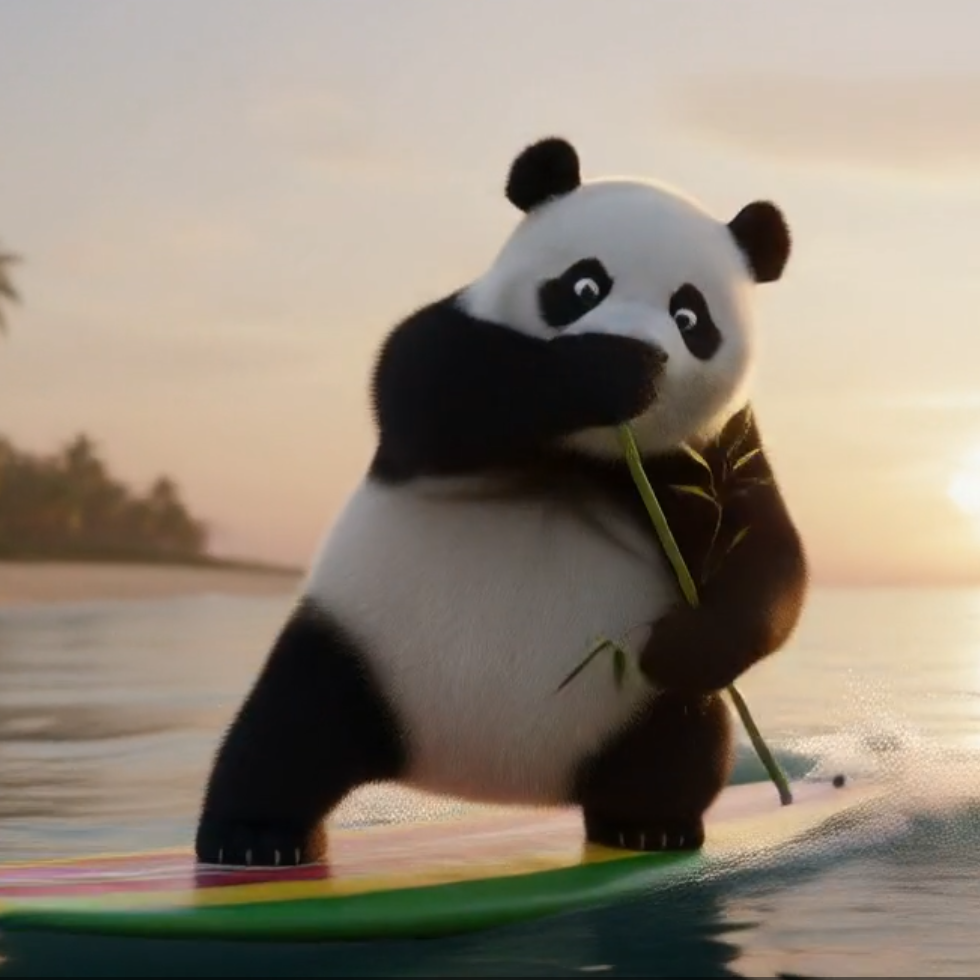} \hfill
        \includegraphics[width=0.19\textwidth]{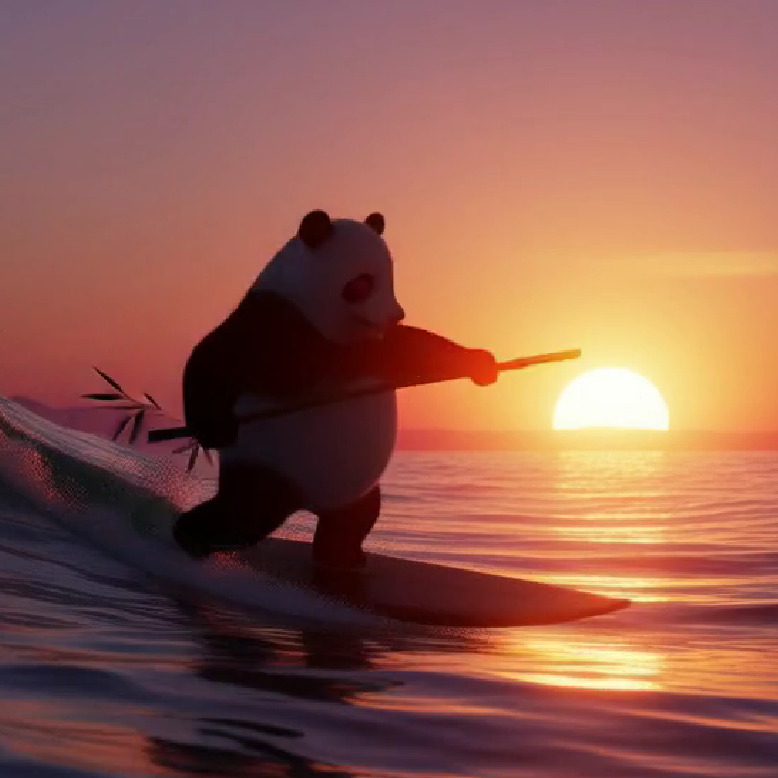}
    \end{minipage}

    \caption{\textbf{Continuous Frame Comparison (Prompt: "A panda standing on a surfboard in the ocean in sunset").} We extract 5 frames from the generated videos. \textbf{Top Row (Sora2):} Shows high fidelity but requires proprietary access. \textbf{Bottom Row (CoAgent):} Our open-source framework maintains robust subject identity (the panda) and consistent lighting interaction with the environment, matching the temporal coherence of the commercial state-of-the-art.}
    \label{fig:sora_comparison}
\end{figure*}

To demonstrate the narrative coherence of CoAgent, we conducted a side-by-side comparison with Sora2 using a dynamic, nature-themed prompt. Figure \ref{fig:sora_comparison} presents a "filmstrip" visualization, extracting 5 continuous frames from generated video sequences to highlight temporal stability.

\paragraph{Narrative Consistency.} 
As illustrated in Figure \ref{fig:sora_comparison}, both models were tasked with generating a sequence based on the prompt: \textit{"A panda standing on a surfboard in the ocean in sunset."}
\begin{itemize}
    \item \textbf{Sora2:} Demonstrates exceptional visual fidelity, particularly in water rendering and lighting reflections.
    \item \textbf{CoAgent (Ours):} Achieves a comparable level of narrative flow. Crucially, our Global Context Manager (GCM) ensures that the panda's specific visual features (e.g., fur texture, body proportions) remain strictly consistent across the motion of surfing, maintaining the subject's identity against the complex, moving background of the ocean waves.
\end{itemize}

\subsection{Limitations: Physical Inconsistencies and Interaction Hallucinations}
\label{subsec:limitations}

\begin{figure}[htbp]
        \centering
        \includegraphics[width=\columnwidth]{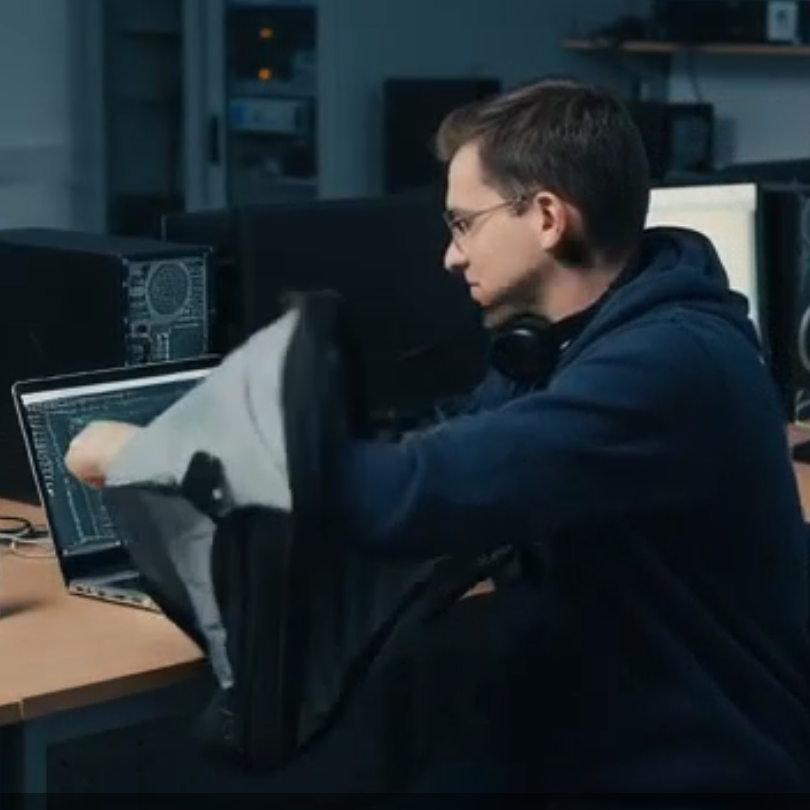} 
        \caption{\textbf{Failure Case: Object Penetration Artifact.} An example of physical hallucination where a hand interacts incorrectly with a laptop bag, clipping through the object's surface. This failure highlights the limitation of 2D-based diffusion backbones in modeling object rigidity, as well as the difficulty for VLM-based verifiers to detect fine-grained, high-frequency physical anomalies due to sparse frame sampling.}

    \label{fig:physics_failure}
    \end{figure}

While CoAgent significantly improves narrative logic and semantic alignment, we acknowledge distinct limitations regarding fine-grained physical interactions. As illustrated in Figure \ref{fig:physics_failure}, the system occasionally produces "physics hallucinations," such as solid objects passing through one another.

\paragraph{Analysis of Failure Modes.}
We attribute these artifacts to two primary factors:

\begin{itemize}
    \item \textbf{Inherited Backbone Priors:} The underlying synthesis model (Wan2.1) operates in 2D pixel space without an explicit 3D physics engine. Consequently, it occasionally fails to model object rigidity and occlusion, leading to artifacts like the "hand clipping through the laptop bag" shown in Figure \ref{fig:physics_failure}. Since CoAgent operates as a high-level orchestration layer, it cannot easily inject low-level physical constraints into the diffusion unet's internal features.
    
    \item \textbf{Verifier Temporal Sparsity:} To maintain computational efficiency, our Verifier Agent employs a \textbf{sparse sampling strategy} (checking keyframes rather than every single frame). Short-duration physical violations—such as a hand momentarily penetrating a surface during a fast motion—may occur \textit{between} sampled frames, effectively bypassing the verification loop. Furthermore, current VLMs excel at semantic checks but lack the precise spatial reasoning to detect subtle intersection artifacts.
\end{itemize}

\paragraph{Future Direction.}
Addressing these limitations will likely require integrating 3D-aware guidance (e.g., depth map constraints) or developing specialized Video Physics Verifiers trained specifically to detect motion anomalies and intersection errors.

\end{document}